\documentclass[journal]{IEEEtran}

\usepackage{cite}
\usepackage[pdftex]{graphicx}
\usepackage{amsmath}
\usepackage{amsthm}
\usepackage{amssymb}
\usepackage{array}
\usepackage{wasysym}
\usepackage{bm}
\usepackage[]{units}
\usepackage{siunitx}
\usepackage{url}
\usepackage{color}
\usepackage{epstopdf}
\usepackage[ruled]{algorithm2e}

\newcommand*\myswitch{{\protect \includegraphics[width=0.8em]{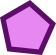}}}
\newcommand*\mypostimpulse{{\protect \includegraphics[width=0.8em]{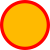}}}

\begin{document}
\title{Robust Dynamic Locomotion via Reinforcement Learning and Novel Whole Body Controller}
% Fully  
% Achieving 
\author{Donghyun~Kim,
        Jaemin~Lee,
 and~Luis~Sentis,~\IEEEmembership{Member,~IEEE}% <-this % stops a space
\thanks{D. Kim and J. Lee are with the Department
of Mechanical Engineering, University of Texas at Austin, Austin, TX, 78712 USA e-mail: .}% <-this % stops a space
\thanks{L. sentis is with the Department of Aerospace Engineering and Applied Mechanics, University of Texas at Austin, Austin, TX, 78712}% <-this % stops a space
\thanks{Manuscript received April 19, 2005; revised August 26, 2015.}}

% The paper headers
\markboth{Journal of \LaTeX\ Class Files,~Vol.~14, No.~8, August~2015}%
{Shell \MakeLowercase{\textit{et al.}}: Bare Demo of IEEEtran.cls for IEEE Journals}

\maketitle

\begin{abstract}
We propose a robust dynamic walking controller consisting of a dynamic locomotion planner, a reinforcement learning process for robustness, and a novel whole-body locomotion controller (WBLC). Previous approaches specify either the position or the timing of steps, however, the proposed locomotion planner simultaneously computes both of these parameters as locomotion outputs. Our locomotion strategy relies on devising a reinforcement learning (RL) approach for robust walking. The learned policy generates multi step walking patterns, and the process is quick enough to be suitable for real-time controls. For learning, we devise an RL strategy that uses a phase space planner (PSP) and a linear inverted pendulum model to make the problem tractable and very fast. Then, the learned policy is used to provide goal-based commands to the WBLC, which calculates the torque commands to be executed in full-humanoid robots. The WBLC combines multiple prioritized tasks and calculates the associated reaction forces based on practical inequality constraints. The novel formulation includes efficient calculation of the time derivatives of various Jacobians. This provides high-fidelity dynamic control of fast motions. More specifically, we compute the time derivative of the Jacobian for various tasks and the Jacobian of the centroidal momentum task by utilizing Lie group operators and operational space dynamics respectively. The integration of RL-PSP and the WBLC provides highly robust, versatile, and practical locomotion including steering while walking and handling push disturbances of up to 520 N during an interval of 0.1 sec. Theoretical and numerical results are tested through a 3D physics-based simulation of the humanoid robot Valkyrie.
\end{abstract}


\IEEEpeerreviewmaketitle

%%%%%%%%%%%%%%%%%%%%%%%%%%%%%%%%%%%%%%%%%%%%%%%%%%%%%%%%%%
%%%                   Introduction                      %%
%%%%%%%%%%%%%%%%%%%%%%%%%%%%%%%%%%%%%%%%%%%%%%%%%%%%%%%%%%
\section{Introduction}

%Full-humanoid walking using Reinforcement Learning (RL) has not been achieved despite attempts to utilize RL on biped locomotion. 
%Reinforcement Learning (RL) based locomotion planners rarely compete with model-based algorithms because learning high-dimensional continuous values is difficult. 
%Moreover, RL frequently leads to a local minimum, particularly when the region to explore is large and the problem is highly nonlinear.
%Instead of exploring an entire state space, we formulated RL in a constrained space by considering the relationships between locomotion parameters.

We explore the performance of a reinforcement learning (RL) process and a new whole-body impedance and force controller for robust dynamic locomotion on a full biped humanoid dynamical model. Full-bodied 3D humanoid dynamic walking based on inverted pendulum (IP) dynamics and RL has not been studied to date. To enable the RL process to run efficiently, we found that utilizing phase-space planning (PSP) \cite{Zhao:2012de} provides a space of practical parameters that enables the transition function to operate in the reduced inverted pendulum manifold. A key advantage of using an IP model is that it generalizes locomotion for many types of systems with a light dependency with their concrete kinematic structure. Using an IP model not only reduces the search space but also enables the same procedure to be used across different types of full-bodied bipedal humanoid robots. A closely related work to ours, \cite{MacAlpine:2012vp} utilizes a trajectory parametrization of the dynamic locomotion process \cite{Graf:2009uz}, specifically customized for a particular robot, the NAO. As such, our study can be viewed as a generalization of this type of work to more general models typically used in the dynamic walking communities \cite{dynamic-walking-2017} -- e.g. inverted pendulums. In addition, the previous line of work on NAO robots, has remained quiet with respect to quantifying robustness. In the work presented here, we address robustness as a main thrust performing detailed simulations with respect to unplanned and large disturbances at moderate walking speeds.  

Since IP dynamics are in a reduced manifold, we propose a new type of whole-body controllers that are highly robust and effective to transfer the IP-based locomotion process into the full humanoid model. In particular we incorporate new efficiently-computed feedforward terms, momentum and balancing tasks, and more accurate contact models while maintaining the key capability of task prioritization. In addition we completely reformulate the control structures for whole-body control with respect to previous work of ours on this area of whole-body control. We accommodate for the new models and also achieve high computational efficiency. As such, we build on our long time history of devising whole-body controllers, this time around making significant algorithmic changes. We believe these transformations constitute a quantum leap with respect to whole-body controllers with dynamic locomotion capabilities. 

The combination of RL-IP-PSP for locomotion pattern generation achieves significant robustness by training a neural network through an actor-critic process with many possible center-of-mass states, representing potential disturbances, then learning successful step timing and foot position policies for recovery. Utilizing step timing and foot positions is not typically explored because simultaneously varying both of these parameters results in nonlinear system dynamics. \cite{Herdt:2010bh} proposed a model predictive control (MPC) method for synthesizing walking patterns based on desired foot positions, kinematic limits, and given step timing. \cite{Kryczka:2015ck} formulated a nonlinear optimization problem to solve two walking steps ahead of time to reduce computational cost. \cite{Khadiv:2016hm} linearized an optimization problem by searching for a solution one step ahead of time.  In contrast, instead of relying on runtime optimizations, we train a control policy offline using an IP model, and use it afterwards for real-time control of full humanoid robots being physically disturbed. Therefore, our learned locomotion planning generator can plan hundreds of steps ahead of time in an instant, compared to the stepping time scale. As such, speed is a key characteristic of the proposed planning and control framework compared to the state of the art. 

Devising a new robust dynamic locomotion generator is insufficient to be directly used in full humanoid robots. Therefore building on our expertise in this area, we devise a new type of whole-body controller (WBC) which we call whole-body locomotion controller (WBLC) that focuses on speed, unilateral contact constraints, and speedy prioritized task control. The proposed WBC algorithm enables to efficiently compute projection-based hierarchical task controllers \cite{Sentis:COl04Q1j} and at the same time incorporate contact inequality constraints which are represented by a quadratic programming (QP) process\cite{koolen2013summary,feng2015optimization,kuindersma2014efficiently}, hierarchical quadratic programming (HQP) \cite{saab2013dynamic}. While QP based controllers have been very successful for field application their computational cost is considerably higher than that of projection-based methods. In contrast, projection-based methods have not incorporated before inequality constraints such as unilateral contact and friction constraints. Our proposed WBC algorithm combines for first time QP and  projection-based methods.

%%%%%%%%%%%%%%% Begin of Valkyrie Figure %%%%%%%%%%%%%%
\begin{figure}
\centering
\includegraphics[width=0.8\columnwidth]{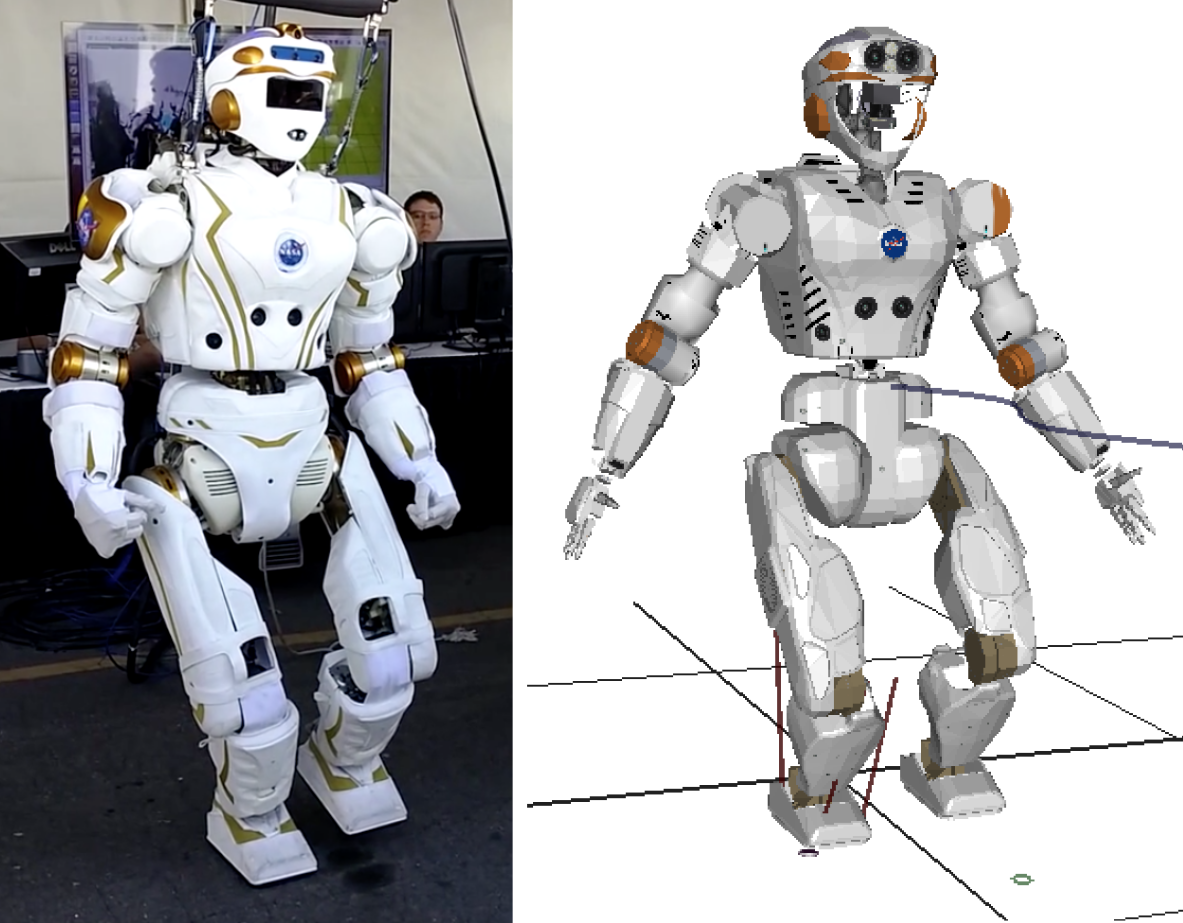}
\caption{{\bf Type of humanoid platform our controller explores.} The left image shows NASA's Valkyrie humanoid robot, with 135.9 $\si[per-mode=symbol]{\kilo\gram}$ weight and 1.83 $\si[per-mode=symbol]{\meter}$ height. The right image shows our dynamic simulation of Valkyrie using the physics based simulator SrLib.}
\label{fig:valkyrie_figure}
\vspace{-1mm}
\end{figure} 
%%%%%%%%%%%%%%% End of Valkyrie Figure %%%%%%%%%%%%%%

In our study, we introduce a centroidal angular momentum (CAM) tasks to improve agile locomotion performance and we examine its effect when used in the whole-body control hierarchy. For example, when a robot quickly shakes or rotates its body, we reduced undesirable arm movements that result from angular momentum compensation by introducing an arm motion task with higher priority than the CAM task. 

In the proposed WBLC, we devise new projection-based recursive structures that incorporate unilateral contact and friction constraints, yielding the desired reduction on the computational cost compared to other QP based algorithms. However, we don't only achieve computational efficiency by combining QP and projection-based methods. We do achieve it by important improvements on the computation of the projection-based operations themselves.

Indeed, in conventional projection-based methods, the computational cost of some operations is considerably high. For instance, one well-known WBC algorithm that uses joint acceleration, \cite{siciliano1991general}, includes costly terms such as the time derivative of a null space projection matrix. In addition, many WBC algorithms contain computations for Coriolis/centrifugal and gravitational forces projected on in operational task space \cite{sentis2005synthesis,mansard2009unified}, which are costly to calculate specially as the number of control tasks increases. Our WBLC algorithms eliminates this problems. An analytic solution of the time derivatives of Jacobians is devised by employing Lie group operators, and an implementation using the Rigid Body Dynamic Library. We also eliminate the need to compute Coriolis/centrifugal terms for every task priority.

%%%%%%%%%%%%% Begin Phase Space Figure %%%%%%%%%%%%%%%
\begin{figure}
\centering
\includegraphics[width=\columnwidth]{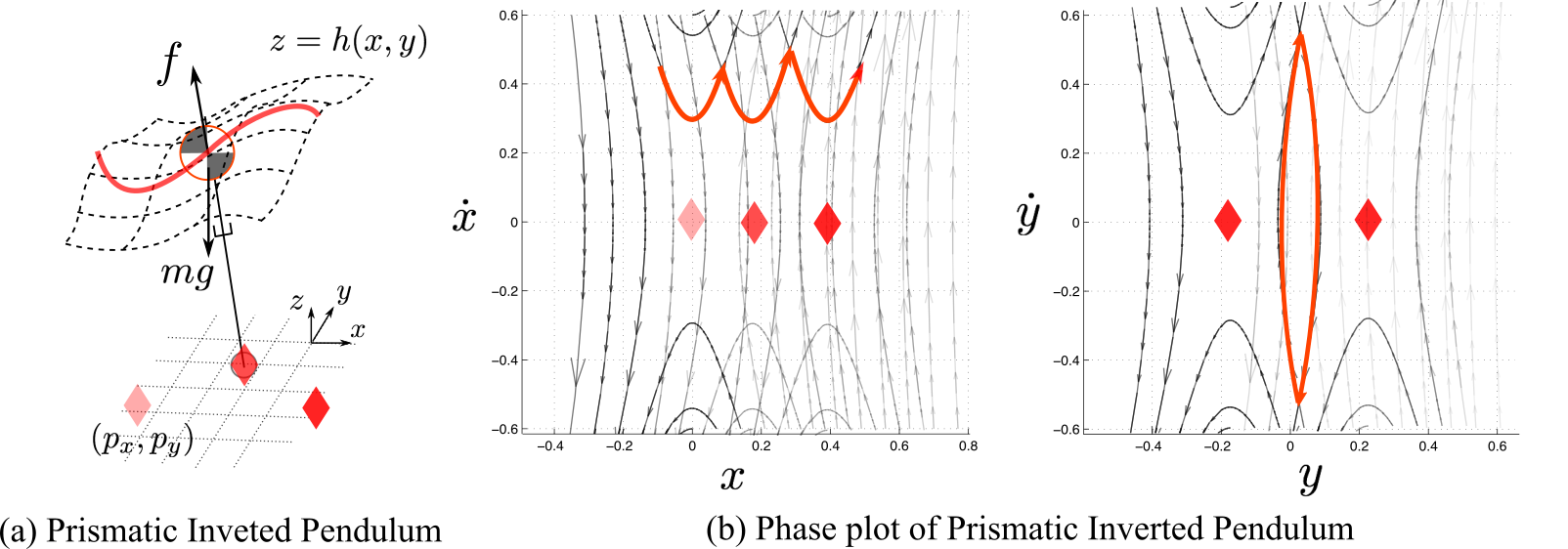}
\caption{{\bf PIPM and CoM phase plots.} (a) PIPM 3D position moving on a variable height surface. (b) Overlapped PIPM phase plots corresponding CoM paths during dynamic walking. In the sagittal plane, we can see multiple parabolas connected to each other corresponding to various walking steps. Parabolas in the frontal plane produce limit cycles.}
\label{fig:psp_concept}
\end{figure}
%%%%%%%%%%%%% End Phase Space Figure %%%%%%%%%%%%%%%

Overall, the main contributions of our study are as follows. First, devising a novel learning framework for robust dynamic locomotion under push disturbances achieving virtually instantaneous re-planning of an order of magnitude more steps that the state-of-the-art. Second, we devise an elegant method to introduce steering capabilities to phase-space planning for dynamically moving in all directions. Third, we devise a new whole-body locomotion controllers, which yields the benefits of QP based computation of reaction forces and projection-based prioritized task control. Due to many optimizations, we believe that this controller is one of the fastest WBC's that fulfills both prioritization and practical inequality constraints. Lastly, we integrate all of these algorithms into a comprehensive software and conduct thorough testing on robust dynamic locomotion under large push disturbances on physics-based simulations of Valkyrie.

%%%%%%%%%%%%%%% Begin of Phase Space Explanation %%%%%%%%%%%%
\begin{figure*}
\centering
\includegraphics[width = 1.95\columnwidth]{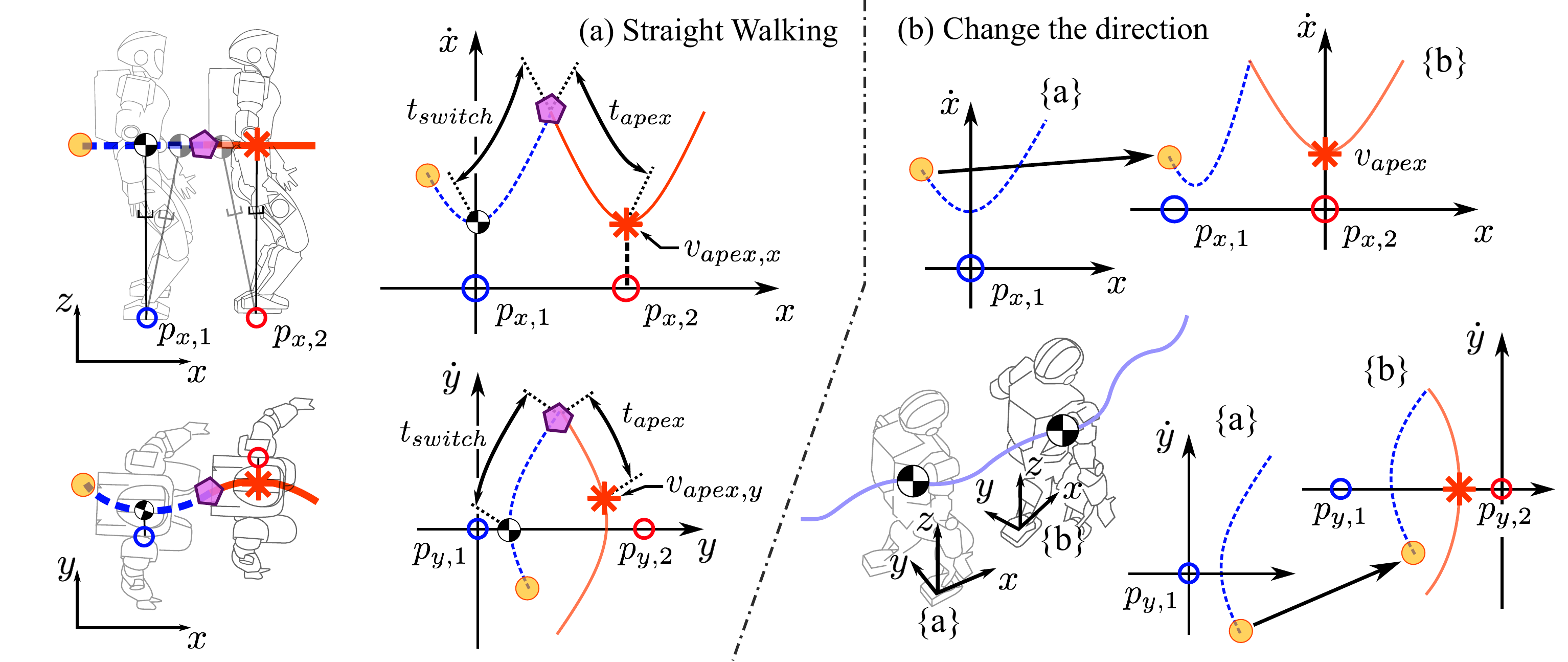}
\caption{{\bf Phase Space Planner (PSP).} (a) shows the method to find a switching time and a lateral foot placement position given forward step location and an apex velocity. In the sagittal phase plot, we can see the given current CoM state and the apex state uniquely define switching states \myswitch. From an initial state, the planner computes the switching and apex times. These two timing values are used to find the next step location in the lateral plane. (b) is the process to steer the robot's walking direction. When changing the walking direction, we first align the orientation of the next local frame to the direction the robot intends to go, and second we project the current CoM state into the next local frame. }
\label{fig:psp_exp}
\end{figure*}
%%%%%%%%%%%%%%%% End of Phase Space Explanation %%%%%%%%%%%%

The paper is organized as follows. In Section~\ref{sec:rl_psp}, we describe the RL-IP-PSP process after briefly reviewing related work in Section~\ref{sec:related_work}. We then introduce the formulation of WBLC in Section~\ref{sec:wblc} and the methods to efficiently obtain the time derivative of Jacobians for motion control and CAM tasks. In Section~\ref{sec:result}, we study the effects of our framework in agile task such as shaking the robot's body, walking while steering, and recovering from large pushes while walking. We do all of this using a model of NASA's Valkyrie robot and the srLib physics based engine\footnote{Seoul National University Robotics Library. Physics-based simulation. Open-source \url{http://robotics.snu.ac.kr/srlib/}}(Fig.\ref{fig:valkyrie_figure}). A more exhaustive review of previous work can be found in Appendix~\ref{sec:related_work}. 

%%%%%%%%%%%%%%%%%%%%%%%%%%%%%%%%%%%%%%%%%%%%%%%%%%%%%%%%%%
%%%      Reinforce Learning based Phase Space Planner   %%
%%%%%%%%%%%%%%%%%%%%%%%%%%%%%%%%%%%%%%%%%%%%%%%%%%%%%%%%%%

\section{Reinforcement Learning based \\Phase Space Planner}
\label{sec:rl_psp}
We devise an RL process around a PSP framework, the latter significantly enhancing the learning efficiency by exploiting the inherent directional walking constraints of PSP. PSP generates effective step switching information using simplified models such as the prismatic inverted pendulum model (PIPM). In Fig.~\ref{fig:psp_concept}, we show phase plots across multiple walking steps of the CoM sagittal and lateral phase portraits based on PIPM dynamics. In the sagittal plane, the path consists of connected parabolas, while in the frontal plane, the walking path follows semi-periodic parabolas in a closed cycle. For convenience, we will use $x$ for the sagittal plane and $y$ for the frontal plane.

\subsection{Phase Space Planner}
\label{sec:psp_explain}

Leading step planning generators, such as Divergent Component of Motion \cite{Englsberger:2015jp}, find CoM paths given step positions and their timing as input information. The ZMP Preview Control method \cite{Kajita:2003iq} has different mechanics but the generated output can be interpreted as finding the CoM path given step position and timing input information. In contrast, PSP finds the step switching time and lateral foot positions, given sagittal foot positions and apex velocities (Fig.~\ref{fig:psp_exp}). The apex states are those at the instant when the sagittal CoM velocity is at its minimum; equivalently, they can be considered as states where the sagittal CoM position is zero in a local frame attached to the stance foot, i.e. below the CoM sagittal position.  

In Fig.~\ref{fig:psp_exp}(a), we can see that the current robot's CoM state and the next desired apex state uniquely define a switching state \myswitch, a switching time, and an apex time. These timings are used to find the next lateral foot position, $p_{y,2}$. Note that the resultant locomotion trajectory is straight forward if $\dot{x}_{apex}$ is a positive number and $\dot{y}_{apex} =0$. In contrast, the proposed algorithm, applies a simple and elegant modification that allows to dynamically steer the biped in any direction of walking (see Fig.~\ref{fig:psp_exp}(b)). When we need to turn walking direction, we re-initialize the orientation of the local frame $\{b\}$ to the new direction and project the current state to the new frame. The original PSP algorithm devised locomotion trajectories via numerical integration. However, for algorithmic speed purposes, the methods presented here assume that the CoM height is linear allowing us to exploit an analytical solution (see Appendix.~\ref{sec:append_psp}).  

% Benefit of PSP
Considering a one step ahead plan, an initial CoM state, and desired future states $[p_x,~\dot{x}_{apex},~\dot{y}_{apex}]^{\top}$ PSP finds the next step position and timing, $\begin{bmatrix} p_y, &t_{switch}\end{bmatrix}^{\top}$. Notice that the walking direction is indicated using apex velocities, $[\dot{x}_{apex}, ~\dot{y}_{apex}]^{\top}$. We will now see that our formulation of PSP makes the RL problem more efficient by reducing the dimensionality of the learned state variables.

\subsection{The Reinforcement Learning Problem}
\label{sec:rl_process}

\begin{figure}
\includegraphics[width=\columnwidth]{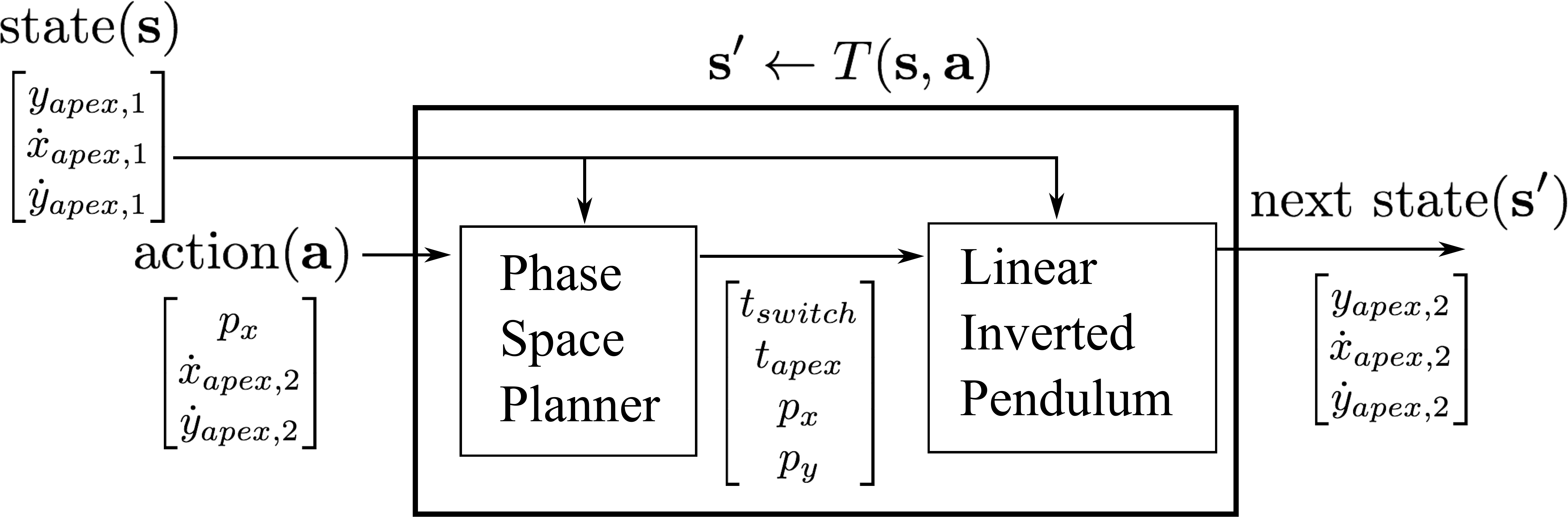}
\caption{{\bf Transition Function.} The transition function relies on two models: PSP and LIPM. Given an apex state and a considered action, PSP computes step timing and location information that serve as inputs to the LIPM model. Then, LIPM solves for the next apex state based on the given state and the provided inputs.}
\label{fig:transition_fn}
\end{figure}

% update rule
As mentioned before, a central part of our walking methodology is to achieve robustness via reinforcement learning. The technique we use is the Actor-Critic with Eligibility Traces method. We summarize this process in Algorithm~\ref{code:rl_explain} which is an adaptation \cite{sutton2011reinforcement}.  
\begin{algorithm}[t]
% Value function
 \KwIn{$\hat{v}(\mathbf{s}, \mathbf{w})$, $\forall \mathbf{s} \in \mathcal{S}$, $\mathbf{w} \in \mathbb{R}^{18 \times 30 \times 56 + 1}$}
% pi (policy)
  \KwIn{$\pi( \mathbf{a}| \mathbf{s}, \bm{\theta})$, $\forall a \in \mathcal{A}$, $\mathbf{s} \in \mathcal{S}$, $\bm{\theta} \in \mathbb{R}^{(18 \times 30 \times 56 + 1) \times 6}$}
 \vspace{1.5mm}
 \KwResult{$\bm{\theta}$, $\mathbf{w}$} \vspace{1.5mm}
 Initialize policy weights $\mathbf{\theta}$ and state-value weights $\mathbf{w}$\vspace{1.5mm}
 \While{(variances of policy is large)}{\vspace{1mm}
	Randomly pick $\mathbf{s}$ in $\mathcal{S}$\\
	$\mathbf{e}^{\bm{\theta}} \gets \mathbf{0}$ (eligibility trace of policy parameters)\\
	$\mathbf{e}^{\mathbf{w}} \gets \mathbf{0}$ (eligibility trace of value parameters)\\
	 $I \gets 1$\\ \vspace{1mm}
	\While{($\mathbf{s}$ is not terminal)}{\vspace{1mm}
	$\mathbf{a} \sim \pi(\cdot | \mathbf{s}, \bf{\theta})$\\
	$\mathbf{s'}, R \gets T( \mathbf{s}, \ \mathbf{a})$\\
	$\delta \gets R + \gamma \hat{v}(\mathbf{s'}, \mathbf{w}) - \hat{v}(\mathbf{s}, \mathbf{w})$ \\
	$\mathbf{e}^{\mathbf{w}} \gets \lambda^{\mathbf{w}}\mathbf{e}^{\mathbf{w}} + I\nabla_{\mathbf{w}} \hat{v}(\mathbf{s}, \mathbf{w})$ \\
	$\mathbf{e}^{\bm{\theta}} \gets \lambda^{\bm{\theta}}\mathbf{e}^{\bm{\theta}} + I\nabla_{\bm{\theta}} \log \pi(\mathbf{a}|\mathbf{s}, \bm{\theta})$ \\
   	$\mathbf{w}\gets \mathbf{w} + \beta \delta \mathbf{e}^{\mathbf{w}}$\\
    $\bm{\theta}\gets \bm{\theta} + \alpha \delta \mathbf{e}^{\bm{\theta}}$\\
	$I \gets \gamma I$ \\
    $\mathbf{s} \gets \mathbf{s'}$\\
	} 
}
 \caption{Actor-Critic with Eligibility Traces}\label{code:rl_explain}
\end{algorithm}
We define $\mathbf{s}$, as CoM apex states,  
$\mathbf{s} \triangleq \begin{bmatrix} y_{apex}, &\dot{x}_{apex}, &\dot{y}_{apex} \end{bmatrix}^{\top}$.
Notice that $\mathbf{s}$ does not include the variable $x_{apex}$ because it is assumed to be always zero in the local frame. We define actions, $\mathbf{a} \triangleq \begin{bmatrix} p_x, &\dot{x}_{apex}, &\dot{y}_{apex} \end{bmatrix}^{\top}$, as input parameters to the PSP process. 
A transition function, $T(\mathbf{s},~\mathbf{a})$, computes the next apex state, $\mathbf{s}'$, and the instantaneous reward value. In Fig.~\ref{fig:transition_fn}, we show the transition function, consisting of two stages: 1) finding step timing and position values via PSP, and 2) computing the next apex state via an analytic solution of the linear inverted pendulum model (LIPM). The first stage is described in Appendix~\ref{sec:append_psp} and allows to find $t_{switch}$, $t_{apex}$, and $p_y$ from the current apex state and chosen action. The second stage, finds the next apex state using the analytic solution of the CoM dynamics (see Eq.~\eqref{eq:x_state}). In Algorithm \ref{code:Phase_Space} the process of finding the switching times and the next apex states is explained in detail.

The next item, $\hat{v}(\mathbf{s}, \bm{w})$, corresponds to the value function \-- similar to the cost-to-go function in Dynamic Programming. We store its learned values using a radial basis function (RBF) neural network~\cite{cualinmultidimensional}. The network uses a three-dimensional input vector consisting of the CoM apex state.
\begin{equation}\label{eq:state_range}
\begin{split}
& \cdot ~ -0.14 \leq y_{apex} \leq 0.2 ~(\si[per-mode=symbol]{\meter}),\\[1mm]
& \cdot ~ \ \  0.03 \leq \dot{x}_{apex} \leq 0.61 ~(\si[per-mode=symbol]{\meter\per\second}), \\[1mm]
& \cdot ~  -0.55 \leq \dot{y}_{apex} \leq 0.55 ~(\si[per-mode=symbol]{\meter\per\second}).
\end{split}
\end{equation}
%
% Note that the state scope for $y_{apex}$ is near to zero since we assume that every step is right stance and left leg swing. It does not mean that the method cannot address a left stance case since we can flip the state in the frontal plane because CoM behavior is symmetric on the plane. 
The hidden layer consists of a bias term and $18\times 30 \times 56$ Gaussian functions with centers on a grid with 2$\si[per-mode=symbol]{\centi\meter}$ spacing along each input dimension. The policy function also consists of an RBF neural network but a little different from the value function because of actions are chosen based on an stochastic evaluation.

%%%%%%%%%%%%%% Begin of Transition Function %%%%%%%%%%%%%%%%%
%%%%%%%%%%%%%% End of Transition Function %%%%%%%%%%%%%%%%%

\begin{figure}
\centering
\includegraphics[width=\columnwidth]{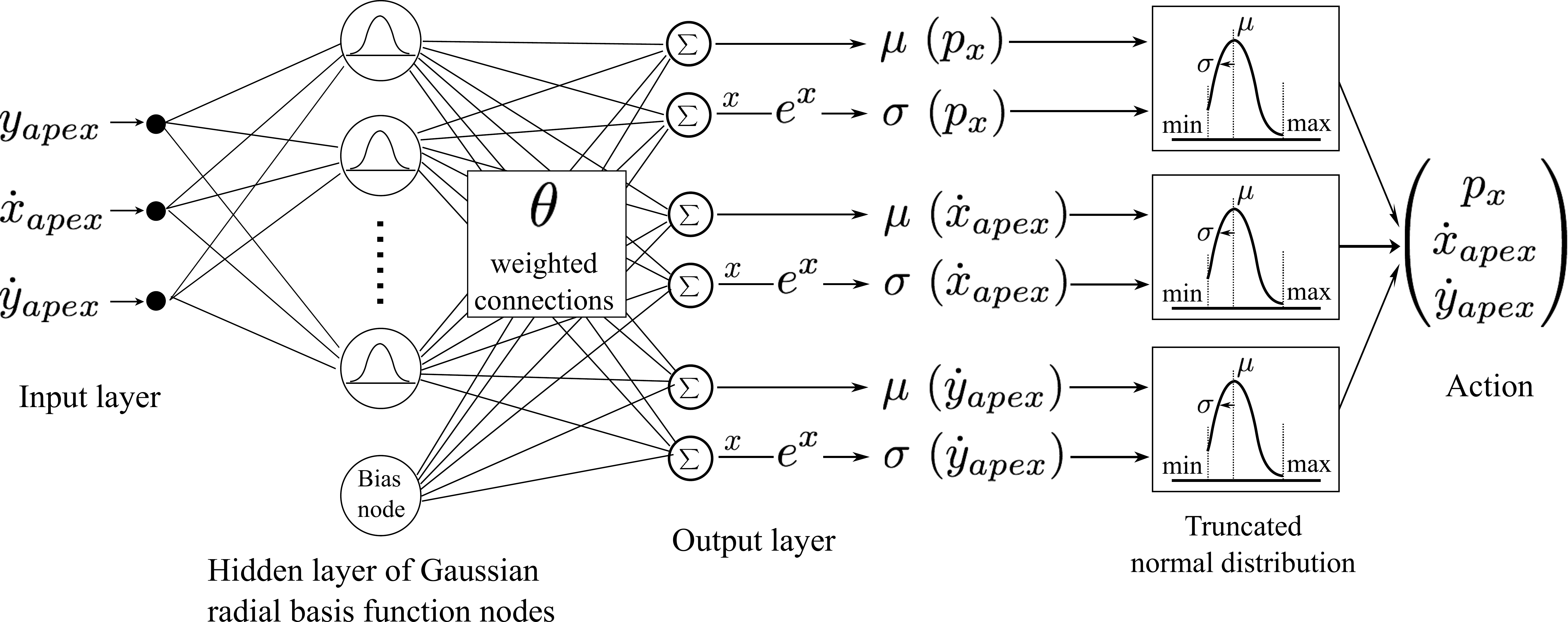}
\caption{ {\bf Radial Basis Function Neural Network for Walking Policy Representation.} Outputs of the neural network are means and standard deviations of each action value. The truncated normal distributions defined by the outputs are used to stochastically pick actions.}
\label{fig:rb_policy}
\end{figure}

Fig.~\ref{fig:rb_policy} shows that outputs of the RBF network are means and standard deviations of truncated normal distributions, $\pi(\mathbf{a}|\mathbf{s}, \bm{\theta})$. The range of the distributions are selected by considering the desired walking speed and step length limits as follows:
\begin{equation}
\begin{split}
& \cdot ~    0.1 \leq p_x \leq 0.5 ~ (\si[per-mode=symbol]{\meter}),\\[1mm]
& \cdot ~    0.03 \leq \dot{x}_{apex} \leq 0.37 ~(\si[per-mode=symbol]{\meter\per\second}),\\[1mm]
& \cdot ~   -0.25 \leq \dot{y}_{apex} \leq 0.25 ~(\si[per-mode=symbol]{\meter\per\second}).
\end{split}
\end{equation}
The network's outputs are linearly weighted by $\bm{\theta}$; thus, the purpose of RL is to find the weights $\bm{\theta}$ given candidate actions that minimized the desired cost.   

%%% Let's define Transition function T(s,a) 
The instantaneous reward is defined by the forward velocity error and lateral step size error:
\begin{equation}
R = -(\dot{x}_{apex}^{nom} - \dot{x}_{apex})^2 - 15 \times (p_y^{nom} - p_y)^2 -(\dot{y}_{apex})^2.
\end{equation}
The set target for the learning process is to achieve recovery behaviors that maintain a straight forward direction, $\dot{y}_{apex} = 0$ while keeping a nominal lateral directional step size. We choose $\dot{x}_{apex}^{nom} = 0.2\si[per-mode=symbol]{\meter\per\second}$ and $p_{y}^{nom} = 0.3\si[per-mode=symbol]{\meter}$. The reward comes from the transition function described before, given current apex and action states selected from the truncated distributions. 

If the next predicted apex state incurs a terminal condition, the transition function gives a negative reward of $-5.0$, and the process terminates and starts a new iteration. The set of safe conditions (i.e. opposite to the terminal conditions) is the intersection of the following predicates:
\begin{equation}\label{eq:terminal_cond}
\begin{split}
&\cdot ~ t_{apex} > 0.12 \ (\si[per-mode=symbol]{\second}),\\[1mm]
&\cdot ~ t_{switch} > 0.12 \ (\si[per-mode=symbol]{\second}),\\[1mm]
&\cdot ~ 0.1 < p_{y} < 0.5 \   (\si[per-mode=symbol]{\meter}),
\end{split}
\end{equation}
which reflect the ability of the robot to swing its legs and the lateral step length. Notice that we do not include a predicate about the sagittal step length because it is already bounded by the allowable action range. The learning process ends when the variance of the learned policy becomes small enough ($<0.07$ units in our case). The usual number of iterations required to complete the learning process is about 30,000, and the process usually takes about 1 $\si[per-mode=symbol]{\minute}$ to compute on a dual-core, 3.0 GHz, Intel i7 processor thanks to the speed of our analytic PSP method. 
%%%%%%%%%%%%%%%%% Policy Network Figure %%%%%%%%%%%%%%%%%%%
%%%%%%%%%%%%%%% End of Policy Network Figure %%%%%%%%%%%%%%%%

\subsection{Evaluation of Learned Policy}

\begin{figure}
\centering
\includegraphics[width = \columnwidth]{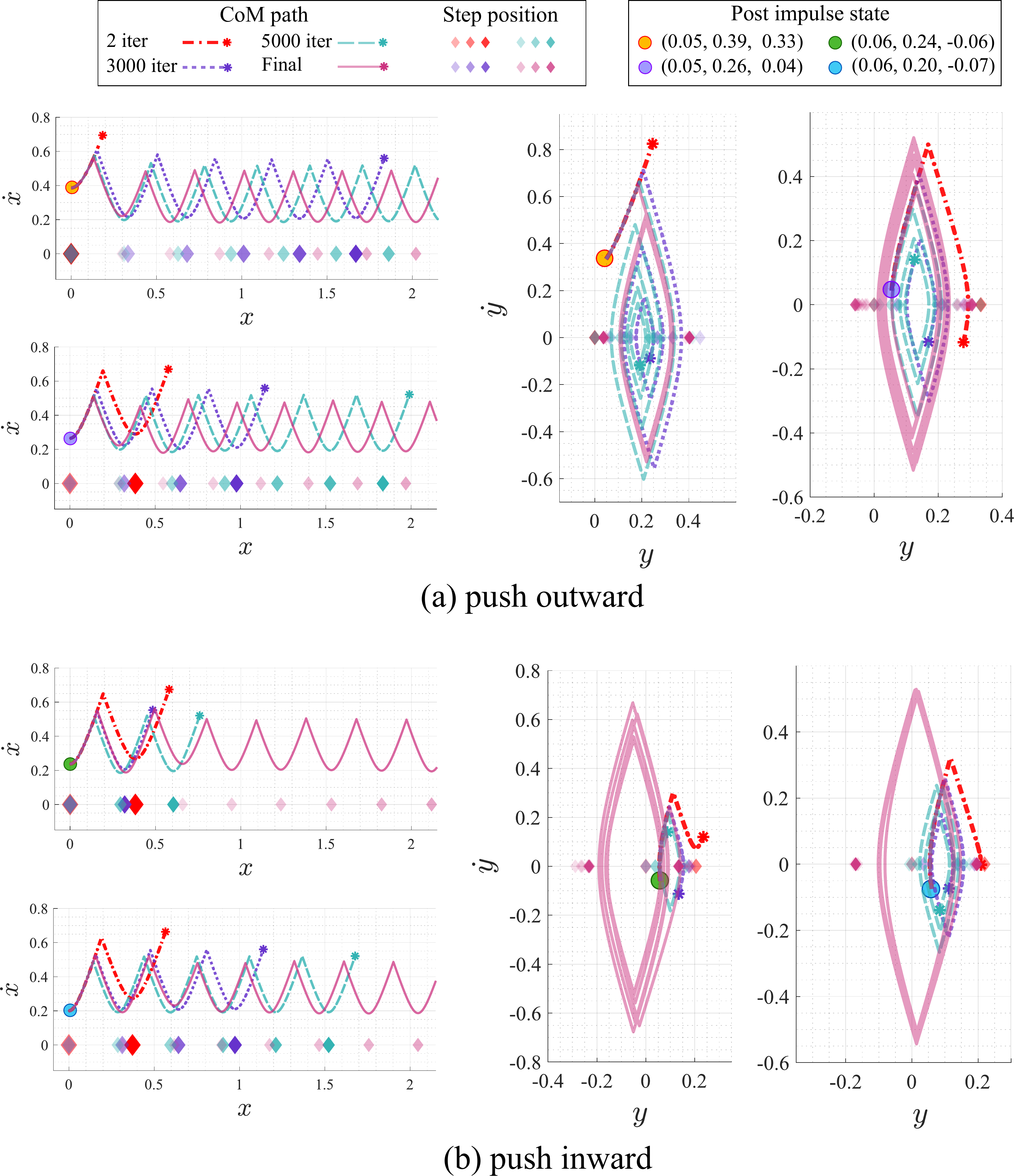}
\caption{{\bf Phase Plots of Sequential Steps from Learned policies.} The initial state considered here corresponds to an impulsive disturbance. The candidate phase trajectories generated by the policy function reach terminal states if they are unsuccesful. As learning proceeds, the policy function finds better actions which avoid terminal states. The final policy achieves infinite number of steps without reaching terminal conditions. }
\label{fig:rl_check}
\end{figure}

Fig.~\ref{fig:rl_check} shows that the performance of the RL-based planner increases as the number of iterations increases. By watching the posture of a robot at different CoM states, we choose the nominal apex state to be $\begin{bmatrix} y_{apex}, &\dot{x}_{apex}, &\dot{y}_{apex} \end{bmatrix}=[~0.056, ~0.2, ~0~]^{\top}$. We proceed by simulating push disturbances to the CoM based on various external forces and directions. We use mean values of the final learned policy as desired actions rather than randomly picking actions from the normal distributions. The results are shown in Fig.~\ref{fig:rl_check} showing the learned policy obtained after many iterations and their enhancements on the walking patterns. In this figure, initially our simulated robot stands with the right foot on the ground and we simulate push disturbances to the left, right, and forward directions of its body. For example, the \mypostimpulse ~post impulse apex state, $[~0.05, ~0.39, ~0.33~]^{\top}$, is the result of an impulse applied to the left-forward direction of the robot's body. Red lines are interrupted within a few walking step indicating that the initial policies fail to find proper actions. In contrast, pink lines correspond to the final learned policy which achieves infinite walking steps without falling given the initial push disturbances. 

%%%%%%%%%%%%%%%%%%%%%%%%%%%%%%%%%%%%%%%%%%%%%%%%%%%%%%%%%%%%%%%%
%%             Whole-Body Locomotion Control                  %%
%%%%%%%%%%%%%%%%%%%%%%%%%%%%%%%%%%%%%%%%%%%%%%%%%%%%%%%%%%%%%%%%

\section{Whole Body Locomotion Control}
\label{sec:wblc}

\begin{figure}
\centering
\includegraphics[width=1.0\columnwidth]{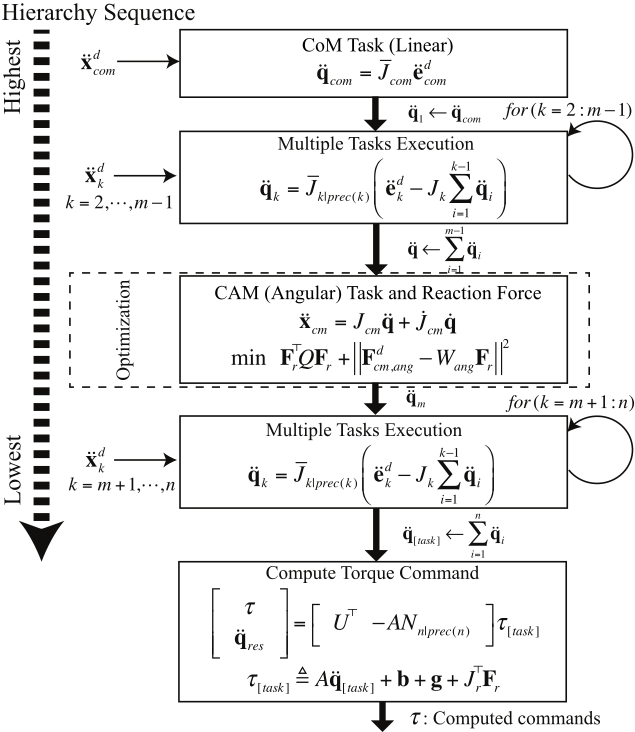}
\caption{{\bf Block Diagram of the Proposed Whole-Body Locomotion Controller.} In WBLC, motion commands are compounded as joint accelerations based on null-space projection methods. The CM task specification is usd to compute reaction forces via QP optimization including unilateral contacts and friction cone constraints. The computed joint acceleration and reaction forces are used to sove for torque commands, which are the final output of WBLC.}
\label{fig:wblc}
\end{figure}

We devise a new whole-body locomotion control algorithm, dubbed WBLC, that specifies tasks using a hierarchy of accelerations and uses quadratic programming to determine contact forces. Fig. \ref{fig:wblc} describes the overall process for computing the torque commands. The details are described below.

%%%%%%%%%%%% Begin of RL Check Result Figure %%%%%%%%%%%
%%%%%%%%%%%% End of RL Check Result Figure %%%%%%%%%%%

\subsection{Acceleration-based Formula with Hierarchy}
Task level controllers are computed in operational space as acceleration commands and converted to joint accelerations using differential forward kinematics,
\begin{equation}
\begin{split}
	\dot{\mathbf{x}}_{1} &= \bm{J}_{1} \dot{\mathbf{q}},\\
	\ddot{\mathbf{x}}_{1} &= \bm{J}_{1}\ddot{\mathbf{q}} + \dot{\bm{J}}_{1}\dot{\mathbf{q}}, 
\end{split}
\end{equation}
where $\mathbf{x} \in \mathbb{R}^{n}$ and $\mathbf{q} \in \mathbb{R}^{m}$ represent the task's operational coordinate and the joint positions, respectively, and $\bm{J}$ is the corresponding Jacobian matrix. Then, the joint acceleration for a desired task acceleration, $\ddot{\mathbf{x}}_{1}^{d}$ can be resolved as
\begin{equation}\label{eq:ddot_q_first}
	\ddot{\mathbf{q}}_{1} = \overline{\bm{J}}_{1} \left( \ddot{\mathbf{x}}_{1}^{d} - \dot{\bm{J}}_{1} \dot{\mathbf{q}} \right) =  \overline{\bm{J}}_{1} \ddot{\mathbf{e}}_{1}^{d},
\end{equation}
where $\overline{\bm{J}}_1$ indicates the dynamically consistent inverse of $\bm{J}_1$, i.e.
\begin{equation}
	\overline{\bm{J}}_{1} = \bm{A}^{-1} \bm{J}_{1}^{\top} \left( \bm{J}_{1} \bm{A}^{-1}\bm{J}_{1}^{\top}  \right)^{-1},
\end{equation}
where $\bm{A} \in \mathbb{R}^{m \times m}$ indicates the mass/inertia matrix of the rigid body model of the robot. If we consider now the mapping of two operational tasks $\ddot{\mathbf{x}}_{1}^{d}$ and $\ddot{\mathbf{x}}_{2}^{d}$, we propose the following task hierarchy mapping
\begin{equation}
	\ddot{\mathbf{q}} = \overline{\bm{J}}_{1} \ddot{\mathbf{e}}_{1}^{d} + \overline{\bm{J}_{2|1}} \left( \ddot{\mathbf{e}}_{2}^{d} -  \bm{J}_{2} \ddot{\mathbf{q}}_{1} \right),
\label{eq:two_task}
\end{equation}
where $\overline{\bm{J}_{2|1}}\triangleq \overline{\left( \bm{J}_{2} \bm{N}_{1} \right)}$ represents the Jacobian associated with the second task, $\bm{J}_2$, projected into the null space of the first task, $\bm{N}_1=\bm{I} -  \overline{\bm{J}}_{1} \bm{J}_{1}$, which by definition is orthogonal to the Jacobian associated with the first task, $\bm{J}_{1}$. The Equation (\ref{eq:two_task}) can be extended to the general $n$ task case, using the following hierarchy
\begin{equation}
	\ddot{\mathbf{q}}_{[task]} =  \overline{\bm{J}}_{1} \ddot{\mathbf{e}}_{1}^{d} + \sum_{k=2}^{n} \ddot{\mathbf{q}}_{k},\quad (n\geq 2)
\label{eq:n_tasks}
\end{equation}
with
\begin{equation}
\begin{split}
	&\ddot{\mathbf{q}}_{k} = \overline{\bm{J}}_{k|prec(k)} \left( \ddot{\mathbf{e}}_{k}^{d} - \bm{J}_{k} \sum_{i=1}^{k-1} \ddot{\mathbf{q}}_{i} \right),\\
    &\bm{J}_{k|prec(k)} = \bm{J}_{k} \bm{N}_{prec(k)}, \\
    &\bm{N}_{prec(k)} = \prod_{s=1}^{k-1} \bm{N}_{s|s-1} \quad (k\geq 2, \quad \bm{N}_{1|0} = \bm{N}_1), \\
&\bm{N}_{s|s-1} = \bm{I} - \overline{\bm{J}}_{s|prec(s)} \bm{J}_{s|prec(s)}
\quad (s \geq 2) \textrm{.}
\end{split}
\end{equation}
This task hierarchy is similar, albeit not identical to \cite{siciliano1991general}. Compared to it, our proposed method is more concise, resulting in less computations for similar control specifications. In particular we do not require the computation of time derivatives of prioritized Jacobians. Details on the similarities and differences between these two works are discussed in Appendix \ref{append_b}.  

%%%%%%%%%%%%%%   Reaction Force Optimization   %%%%%%%%%%%%%%
\subsection{Optimizing Reaction Forces of Underactuated Robots}
Based on the desired joint acceleration given in (Eq.~\eqref{eq:n_tasks}), WBLC finds torque commands via the following equation:
\begin{equation} \label{eq:multi_dyn}
    \bm{A}(\mathbf{q}) \ddot{\mathbf{q}}^{d} + \mathbf{b}(\mathbf{q},\dot{\mathbf{q}}) + \mathbf{g} (\mathbf{q}) + \bm{J}_{r}^{\top}\mathbf{F}_{r} = \bm{U}^{\top} \bm{\tau},
\end{equation}
where $\mathbf{q} \in \mathbb{R}^{n+6}$ and $\mathbf{b}(\mathbf{q},\dot{\mathbf{q}})$ and $\mathbf{g}(\mathbf{q})$ are the  joint space coriolis/centrifugal and gravity terms, respectively. $\mathbf{F}_{r}$ and $\bm{J}_{r}$ represent the reaction forces and the corresponding contact Jacobian. $\bm{\tau} \in \mathbb{R}^{n}$ and $\bm{U}^{\top} \in \mathbb{R}^{(n+6) \times (n)}$ represent the actuator torque commands and the selection matrix mapping actuated torque to the floating base dynamics. Note that $\ddot{\mathbf{q}}^d$ is chosen as,
\begin{equation}\label{eq:qqd}
\ddot{\mathbf{q}}^d = \ddot{\mathbf{q}}_{[task]} + \bm{N}_{n|prec(n)} \ddot{\mathbf{q}}_{res},
\end{equation}
with
\begin{equation}
    \bm{N}_{n|prec(n)} = \bm{N}_{prec(n)} \bm{N}_{n|n-1}
\end{equation}
where $\ddot{\mathbf{q}}_{res}$ is a residual joint acceleration command.  

To find the reaction forces $\mathbf{F}_r$, we specify a centroidal momentum (CM) operational task. A CM task consists of linear and angular momenta portions. The linear part, corresponds to the robot's CoM behavior, $\mathbf{F}_{cm,lin}$, and is typically used for locomotion planning. On the other hand, the angular behavior, the so-called CAM, $\mathbf{F}_{cm,ang}$, is typically set to zero value. Setting the angular task to zero creates conflict with other tasks, such as body rotational tasks. We circumvent this problem by projecting angular behavior as a lower priority task than body rotational tasks as we will soon see. In addition, sometimes it is not possible to simultaneously fulfill linear and angular momentum specifications. For that reason, we specific CoM behavior as a hard constraint while relaxing angular behavior, i.e.
\begin{equation}
\begin{split}
	\min_{\mathbf{F}_r}\quad &   \mathbf{F}_{r}^{\top} \bm{Q} \mathbf{F}_{r} + \| \mathbf{F}_{cm,ang}^d - \bm{W}_{ang} \mathbf{F}_{r} \|^{2} \\[1.5mm]
    \textrm{Subject to.} \quad& \mu |\mathbf{F}_{r,z} | \geq |\mathbf{F}_{r,x}| \\
    & \mu |\mathbf{F}_{r,z} | \geq |\mathbf{F}_{r,y}| \\
    & \mathbf{F}_{cm,lin}^{d} - \bm{W}_{lin}\mathbf{F}_{r} = \mathbf{0}
\end{split}
\end{equation}
where  $\mathbf{F}_{cm,lin}^{d}$ and $\mathbf{F}_{cm,ang}^{d}$ are the desired linear and angular parts of the CM, $\mu$ represents a friction coefficient related to the contact surfaces, $\bm{Q}$ is a weighting matrix, and $\bm{W}_{ang}$ and $\bm{W}_{lin}$ are mappings from reaction forces to angular and linear momenta behaviors. Based on the results of this optimization, $\mathbf{F}_{r}$, the desired value of the CAM task can be calculated as follows:
\begin{equation}\label{eq:cm-definition}
    \bm{I}_{cm} \ddot{\mathbf{x}}_{cm}^{d} = \mathbf{F}_{cm}^{d} = \left[ \begin{array} {cc} \mathbf{F}_{cm,lin}^{d} & \bm{W}_{ang} \mathbf{F}_{r} \end{array} \right]^{\top},
\end{equation}
where $\bm{I}_{cm}$ is a spatial inertial term. Notice that the term $\bm{W}_{ang} \mathbf{F}_{r}$ might be different than $\mathbf{F}_{cm,ang}^{d}$ since the desired angular behavior might violate friction cone constraints. From the above equation, we extract the desired CM acceleration command $\ddot{\mathbf{x}}_{cm}^{d}$ for usage in the controller hierarchy. More concretely, $\ddot{\mathbf{x}}_{cm}^{d} = \left ( \ddot{\mathbf{x}}_{CoM}^{d} \,  \mathbf{\alpha}_{ang}^{d} \right)$, where the first term within the parenthesis is the desired CoM acceleration and the second term is the desired angular acceleration. Both of these commands are used separately in the hierarchy defined in Eq. \eqref{eq:n_tasks}, to produce the joint acceleration command $\ddot{\mathbf{q}}_{[task]}$ which in turn yields $\ddot{\mathbf{q}}^d$ via Eq.~\eqref{eq:qqd}. Plugging this last term into Eq.~\eqref{eq:multi_dyn} we obtain
\begin{equation}
    \bm{A}\left( \ddot{\mathbf{q}}_{[task]} + \bm{N}_{n|prec(n)} \ddot{\mathbf{q}}_{res} \right) +\mathbf{b} + \mathbf{g} + \bm{J}_{r}^{\top}\mathbf{F}_{r} = \bm{U}^{\top} \bm{\tau},
\end{equation}
which can be written in matrix form as
\begin{equation}
\label{eq:final_step}
\left[\begin{array}{cc}
    \bm{U}^{\top} & -\bm{A}\bm{N}_{n|prec(n)}
\end{array}\right]
\left[ \begin{array}{c}
    \bm{\tau} \\ \ddot{\mathbf{q}}_{res}
\end{array}\right] = \bm{\tau}_{[task]},
\end{equation}
where we have defined the term
\begin{equation}
\label{eq:task_torque}
\bm{\tau}_{[task]} \triangleq \bm{A}\ddot{\mathbf{q}}_{[task]} + \mathbf{b} + \mathbf{g} + \bm{J}_r^{\top}\mathbf{F}_r.
\end{equation}
We now have an underdetermined matrix system which can be solved via pseudo inversion as
\begin{equation}
\left[ \begin{array}{c}
    \bm{\tau} \\ \ddot{\mathbf{q}}_{res}
\end{array}\right] = \left[\begin{array}{cc}
    \bm{U}^{\top} & -\bm{A}\bm{N}_{n|prec(n)}
\end{array}\right]^{+} \bm{\tau}_{[task]}
\label{eq:final_cmd}
\end{equation}
where $(.)^{+}$ represents the Moore-Penrose pseudo inverse operation. 

\section{Time Derivative of Jacobian}
\label{sec:jdot}
\begin{figure}
\centering
\includegraphics[width=\columnwidth]{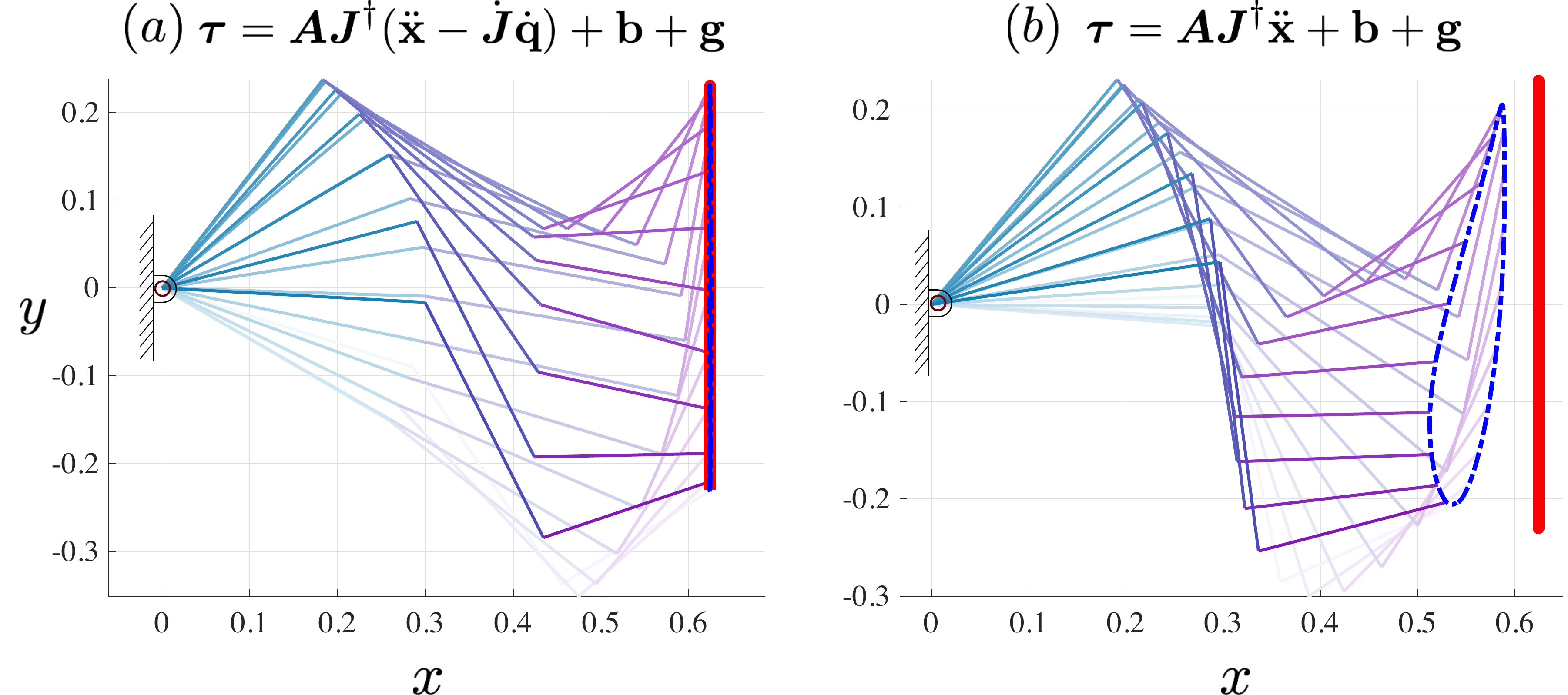}
\caption{{\bf Tracking performance comparison with and without the term $\dot{\bm{J}}\dot{\mathbf{q}}$.} A three-DoF planar manipulator is used to control its end effector to follow a vertical line (red lines) with 2 Hz frequency (blue dashed lines are the end-effector path). The tracking results demonstrte that the (a) controller, which accounts for $\dot{\bm{J}}\dot{\mathbf{q}}$, outperforms the (b) controller.}
\label{fig:jdot_linkage}
\end{figure}
The ability to compute efficiently the time derivative of Jacobian operators 
for fast operational space control has been overlooked. However it plays an important role on robustifying fast movements. Fig.\ref{fig:jdot_linkage} shows that the tracking performance of a simple serial manipulator is significantly enhanced by using the term $\dot{\bm{J}}\dot{\mathbf{q}}$ via Operational Space Control (OPC), where $J$ is the Jacobian of the end effector and $\dot q$ is the vector of joint velocities. Notice that $\dot J$ is used in our WBLC in Equation~\eqref{eq:ddot_q_first}. The commanded task is to follow a vertical line defined by the function, $\mathbf{x}^{d} = [~0.62, ~0.23 \sin(4\pi t)~]^{\top}$. The OPC
input is
\begin{equation}
\ddot{\mathbf{x}} = 
\begin{bmatrix}
0 \\
-36.32 \sin(4\pi t)
\end{bmatrix} + K_p (\mathbf{x}^{d} - \mathbf{x}) + K_v(\dot{\mathbf{x}}^{d} - \dot{\mathbf{x}}).
\end{equation}
We will use Lie group theory to compute the derivatives of point task Jacobians~\cite{kimlie}. We implement this functionality using the Rigid Body Dynamics Library \footnote{Open-source
  \url{https://rbdl.bitbucket.io} }, which is a popular open source dynamics toolbox. In addition, we also devise a new method to compute the time derivative of the CM Jacobian which cannot be computed using Lie group theory.

\subsection{Time Derivative of Point Jacobian}
\label{sec:time_der_pt_jacobian}
\begin{figure}
  \centering
  \includegraphics[width=0.8\columnwidth]{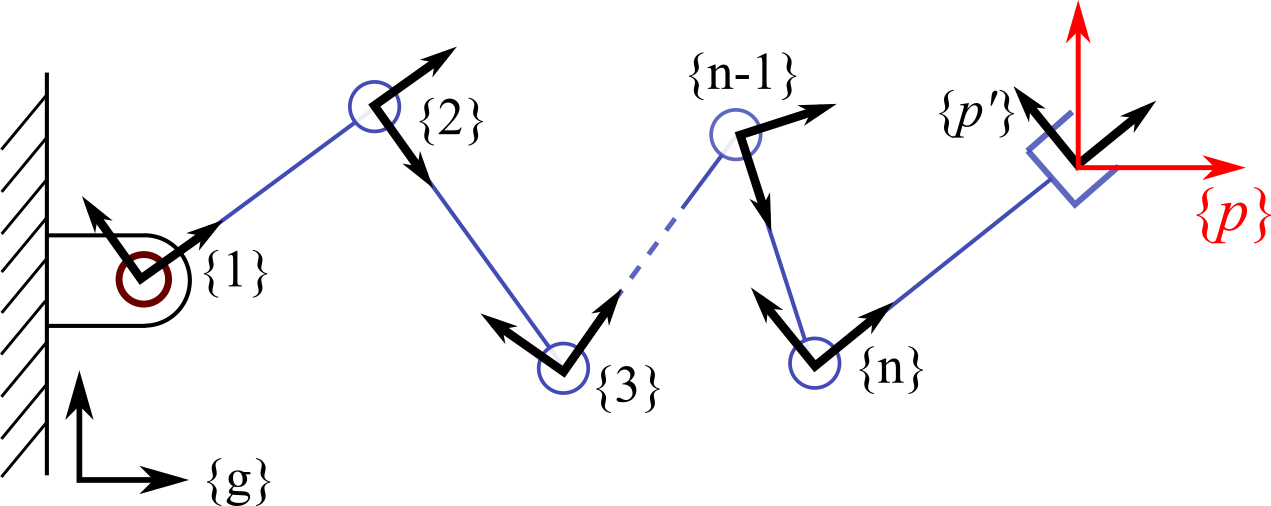}
  \caption[Multi-DoF Openchain]{{\bf Multi-DoF Openchain.} The openchain
    consists of $n$ joints. At the end of the chain, the end-effector is
    attached to link $n$.}
  \label{fig:openchain}
\end{figure}

Lie group operators provide convenient analytic derivations for Jacobian computations. The $SE(3)$ orientation and position representation of a rigid body in three-dimensional space consists of orientation matrix ($\bm{R}$) and a position vector ($\mathbf{p}$). It can also be represented via the $4\times 4$ homogeneous transformation, 
    \begin{equation}
      \bm{T}_{g,i} = \left[
        \begin{array}{cc}
          \bm{R}_{g,i} & \mathbf{p}_{g,i} \\
          0 & 1
        \end{array}
      \right],
    \end{equation}
where, $\bm{R}_{g,i}$ and $\mathbf{p}_{g,i}$ represent the orientation and position of the $i^{th}$ frame in global coordinates respectively (See Fig.~\ref{fig:openchain}). The velocity representation in $se(3)$ consists of the 6-dimensional vector, $[\mathbf{w}, \mathbf{v} ]^T$, and yields the $4\times 4$ homogeneous equality,
    \begin{equation}
      \bm{V}_i \triangleq \left[
        \begin{array}{cc}
          [\mathbf{w}_i]^{\times} & \mathbf{v}_i \\
          0 & 0
        \end{array}
      \right],
    \end{equation}
where 
\begin{equation}
[\mathbf{w}_i]^{\times} \triangleq \begin{bmatrix}
      0 & -w_{i,3} & w_{i,2} \\
      w_{i,3} & 0 & -w_{i,1} \\
      -w_{i,2} & w_{i,1} & 0
        \end{bmatrix}
\end{equation}
Here, $w_{i,1}, w_{i,2}, w_{i,3}$ are relative angular velocities along the three Cartesian axes, and $\mathbf{v}_i$ is the linear velocity. It can be shown that $\bm{V}_i =\bm{T}_{g, i}^{-1} \dot{\bm{T}}_{g,i}$, and corresponds to the generalized velocity seen from the $i^{th}$ frame. The velocity in the global frame associated with $\bm{V}_i$ can be obtained via adjoint derivations, 

\begin{equation}
\begin{split}
{\rm Ad}_{{T}_{g,i}} \left( \bm{V}_i\right) &= \bm{T}_{g,i} \bm{V}_i \bm{T}_{g,i}^{-1} \\
                &= \bm{T}_{g,i} \bm{T}_{g,i}^{-1} \dot{\bm{T}}_{g,i} \bm{T}_{g,i}^{-1}\\
                & = \dot{\bm{T}}_{g,i} \bm{T}_{g,i}^{-1}.
\end{split}
\end{equation}
The adjoint mapping operator is defined as
\begin{equation}\label{eq:adjoint_mtx}
{\rm Ad}_{T_{i,j}} \triangleq  \begin{bmatrix}
\bm{R}_{i,j} & 0 \\
[\mathbf{p}_{i,j}]^{\times} \bm{R}_{i,j} & \bm{R}_{i,j} 
\end{bmatrix},
\end{equation}
where $\bm{R}_{i,j}$ and $\mathbf{p}_{i,j}$ are relative rotations and positions between points $i$ and $j$. The generalized velocity of point $p$ in local coordinates (see Fig.~\ref{fig:openchain}) can be represented as
\begin{equation}
\begin{split}
\mathbf{V}_p & = {\rm Ad}_{{T}_{p,n}} \mathbf{V}_n \\
    &= {\rm Ad}_{{T}_{p,n-1}}\mathbf{V}_{n-1} + {\rm Ad}_{{T}_{p,n}}\mathbf{S}_n \dot{\mathbf{q}}_n \\
    & \qquad \vdots \\
    &= {\rm Ad}_{{T}_{p,0}}\mathbf{V}_{0} + \sum_{i=1}^{n} {\rm Ad}_{{T}_{p,i}}\mathbf{S}_{i} \dot{\mathbf{q}}_{i}.
\end{split}
\end{equation}

%%%%%%%%%%%%%% Body Orientation Control Test %%%%%%%%%%%%%%
\begin{figure*}
\centering
\includegraphics[width=\linewidth]{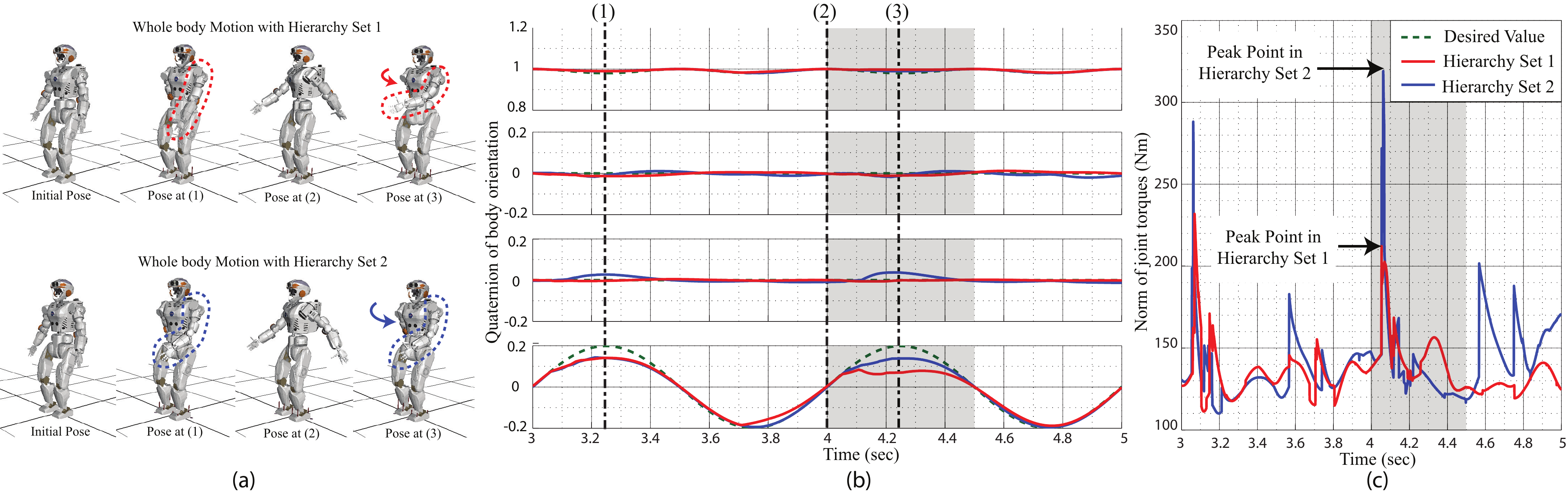}
\caption{{\bf{Body shaking test using different task hierarchies}}: Whole body movements generated via WBLC change corresponding on the task hiearchy. (a) contains simulations using two task hierarchy sets (b) shows data of quaternion tracking error for a body orientation task (c) shows the norm torques generated by WBLC in each case}
\label{fig:shaking_body}
\end{figure*}
%%%%%%%%%%%%%% End of Body Orientation Control Test %%%%%%%%%%%%%%

Because $\{0\}$ frame is the global frame (an inertial frame), $\mathbf{V}_{0}$ is equal to zero. On the other hand $\mathbf{S}_{i}$ maps joint velocities to $\mathit{R}^{6}$, e.g. $\mathbf{S}_i$ is $[~0, 0, 1, 0, 0, 0~]^T$ means $\dot q_i$ is a revolute joint rotating along the $z$ local axis. The first three positions of $\mathbf{S}_{i}$ represent rotational axes while the last three positions represent prismatic axes. It can be shown that the Jacobian of a point $p$, is equal to
\begin{equation}
\bm{J}_p = 
\begin{bmatrix}
{\rm Ad}_{{T}_{p,1}} \mathbf{S}_1
& Ad_{{T}_{p,2}}\mathbf{S}_2  
&\cdots & {\rm Ad}_{{T}_{p,n}}\mathbf{S}_n
\end{bmatrix}
\end{equation}
Furthermore, let's break down the adjoint operators into the following chain operation
    \begin{equation}
      {\rm Ad}_{{T}_{p,i}} \mathbf{S}_i = {\rm Ad}_{{T}_{p, p'}} {\rm Ad}_{{T}_{p', n}} {\rm Ad}_{{T}_{n, i}} \mathbf{S}_i.
    \end{equation}
Here $p'$ is a virtual point representing the position of $p$ but with local orientation with respect to frame $n$. As such it represents just a position offset. In this case $\bm{T}_{p', n}$ is constant and the $i$-th column of the time derivative of the Jacobian can be resolved as
\begin{equation}
\begin{split}
  \dot{\bm{J}}_{p,i}
  & = \dot{\overbrace{\left\{{\rm Ad}_{{T}_{p,i}}\mathbf{S}_i \right\}}} \\
& = \dot{\overbrace{\left\{{\rm Ad}_{{T}_{p, p'}}\right\}}} {\rm Ad}_{{T}_{p', n}} {\rm Ad}_{{T}_{n, i}} \mathbf{S}_i \\
& \quad + {\rm Ad}_{{T}_{p, p'}} {\rm Ad}_{{T}_{p', n}} \dot{\overbrace{\left\{ {\rm Ad}_{{T}_{n, i}} \mathbf{S}_{i} \right\}}} \\
 &= {\rm Ad}_{{T}_{p,p'}} {\rm ad}_{{V}_{p,p'}} {\rm Ad}_{{T}_{p', n}} {\rm Ad}_{{T}_{n, i}} \mathbf{S}_i \\ 
&\quad  + {\rm Ad}_{{T}_{p,p'}} {\rm Ad}_{{T}_{p',n}} \left\{{\rm Ad}_{{T}_{n,i}}  {\rm ad}_{{V}_{n,i}} \mathbf{S}_i + {\rm Ad}_{{T}_{n,i}} (\dot{\mathbf{S}}_i ) \right\}
\end{split}
\end{equation}

Here we have used $\dot{{\rm Ad}_{{T}}} = {\rm Ad}_{{T}} ({\rm
ad}_{\bm{V}})$ and since $V = \bm{T}^{-1}\dot{\bm{T}}$, then we define ${\rm ad}_{\bm{T}^{-1}\dot{\bm{T}}} \triangleq \bm{\bm{T}^{-1}\dot{\bm{T}}} - \bm{\dot{\bm{T}}\bm{T}^{-1}}$. 
% Since ${T}_{p,p'}$ has only rational transformation, ${V}_{p,p'}$ is the angular velocity of $n$ link in its local frame. Instead, ${V}_{n,i}$ is the full body velocity (six dimension) of $n$ link in the local frame. $\dot{\mathbf{S}}_i$ is zero in most cases except the local joint axis is the function of configuration (in our case, every $\mathbf{S}_i$ is constant).

\subsection{Time Derivative of the Centroidal Momentum Jacobian}

The previous equations for the time derivative of point Jacobians are not applicable to the CM Jacobian. The latter can be obtained from the CM task definition of Eq. \eqref{eq:cm-definition} linear part is simply the weighted sum of time derivatives of each link's CoM Jacobian. However, the angular part is not straightforward. Instead of finding $\dot{\bm{J}}_{cm}$, we can find the multiplication of $\dot{\bm{J}}_{cm}$ and the joint velocities, $\dot{\mathbf{q}}$, via operational space dynamics:
\begin{equation}
\bm{\mathit{\Lambda}}_{cm}(\mathbf{q})\ddot{\mathbf{x}} + \bm{\mu}_{cm}(\mathbf{q}, \dot{\mathbf{q}}) + \mathbf{p}_{cm}(\mathbf{q}) = \mathbf{F}_r,
\end{equation}
% where
% \begin{equation}
% \begin{split}
% & \bm{\mathit{\Lambda}}_{cm} = (\bm{J}_{cm}\bm{A}\bm{J}_{cm}^{\top})^{-1},\\
% & \bm{\mu}_{cm} =\bm{\mathit{\Lambda}}_{cm} (\bm{J}_{cm}\bm{A}^{-1}\mathbf{b} - \dot{\bm{J}}_{cm}\dot{\mathbf{q}}), \\
% & \mathbf{p}_{cm} = \bm{\mathit{\Lambda}}\bm{J}_{cm}\bm{A}^{-1}\mathbf{g}.
% \end{split}
% \end{equation}
Here, $\bm{\mathit{\Lambda}}_{cm}$, $\bm{\mu}_{cm}$, and $\mathbf{p}_{cm}$ are
an inertia matrix, coriolis and centrifugal force, and gravitational force of the CM operational task, respectively. Since there is no coriolis and centrifugal effects in CM space, $\bm{\mu}_{cm}$ is zero. Thus,
$\dot{\bm{J}}_{cm}\dot{\mathbf{q}}$ must be equal to
$\bm{J}_{cm}\bm{A}^{-1}\mathbf{b}$:
\begin{equation}
 \dot{\bm{J}}_{cm}\dot{\mathbf{q}} = 
\bm{J}_{cm}\bm{A}^{-1}\mathbf{b}.
\end{equation}
All terms in $\bm{J}_{cm}\bm{A}^{-1}\mathbf{b}$ are easily computable usingoff-the-shelf dynamics libraries. 

\section{Results}
\label{sec:result}

To verify the performance of the proposed methods, we conduct three
demonstrations: 1) body orientation control while changing the task hierarchy, 2) dynamic locomotion with directional change, and 3) push-recovery from various directions while walking. Toward these investigations, we implement our algorithms on a simulation of the Valkyrie humanoid robot, and test it using the physics based simulation SrLib. Because our focus is on locomotion, we fix the finger and wrist joints, bringing the total number of joints to 28. To incorporate floating body dynamics, prismatic and ball joints are introduced to connect Valkyrie's pelvis to a fixed frame.

In the simulation environment, we use a friction coefficient between the ground and the robot's feet of 0.8. On the other hand the friction cone constraints used in WBLC are set to a value of 0.65 to be conservative. In case our contact control solver fails to find proper reaction forces, we allow for solutions that violate friction constraints by relaxing the friction coefficient to a value of 1.75. The resulting control solution implies that slip occurs but only for very short times (in general less than 0.005 $\si[per-mode=symbol]{\second}$). This simple technique doesn't incur an increase in computational complexity while greatly enhancing the robustness of WBLC with respect to external disturbances. 

\subsection{Body Orientation Control with Various Task Hierarchies}
Body shaking behavior is a difficult skill that we use here to study the dynamic performance of WBLC tasks. In traditional humanoid control methods, CoM and CAM tasks are controlled within the same priority level. We propose to split them via WBLC into different hierarchy levels. To demonstrate this feature, we define the following six tasks:
\begin{itemize}
\item[$\cdot$] $\ddot{\mathbf{x}}_{1} \in \mathbb{R}^{3}$: Linear CoM
\item[$\cdot$] $\ddot{\mathbf{x}}_{2} \in \mathbb{R}^{3}$: Centroidal Angular Momentum (CAM)
\item[$\cdot$] $\ddot{\mathbf{x}}_{3} \in \mathbb{R}^{3}$: Body Orientation
\item[$\cdot$] $\ddot{\mathbf{x}}_{4} \in \mathbb{R}^{22}$: Partial Joint Posture (all joints except shoulder pitch, shoulder roll, and knee pitch)
\item[$\cdot$] $\ddot{\mathbf{x}}_{5} \in \mathbb{R}^{3}$: Pelvis Orientation
\item[$\cdot$] $\ddot{\mathbf{x}}_{6} \in \mathbb{R}^{28}$: Full Joint Posture 
\end{itemize}
Most tasks above are self-explanatory. We introduce a partial joint posture consisting on keeping the initial joint positions for all robot joints except for the shoulder pitch and roll, and the knee pitch. This task is used for the sole purpose of testing performance when multiple tasks conflict. In particular, the partial joint posture conflicts with the CAM task within the above task set and viceversa. For our test, we use two hierarchies:
\begin{equation}
\begin{split}
\mathbb{H}_{1} &= \left\{ \ddot{\mathbf{x}}_{1} \rightarrow \ddot{\mathbf{x}}_{2} \rightarrow \ddot{\mathbf{x}}_{3} \rightarrow \ddot{\mathbf{x}}_{4} \rightarrow \ddot{\mathbf{x}}_{5} \rightarrow  \ddot{\mathbf{x}}_{6} \right\} \\
\mathbb{H}_{2} &= \left\{ \ddot{\mathbf{x}}_{1} \rightarrow \ddot{\mathbf{x}}_{3} \rightarrow \ddot{\mathbf{x}}_{4} \rightarrow \ddot{\mathbf{x}}_{5} \rightarrow \ddot{\mathbf{x}}_{2} \rightarrow  \ddot{\mathbf{x}}_{6} \right\}
\end{split}
\end{equation}
The second hierarchy, $\mathbb{H}_{2}$, is more appropriate than the first one, $\mathbb{H}_{1}$ to achieve accurate control of the body shaking (orientation) task. This is accomplished by assigning higher priority to the body orientation task and moving backwards the CAM task. As shown in Fig. \ref{fig:shaking_body} (a), changing the hierarchy levels cause different whole body motions. Fig. \ref{fig:shaking_body} (b) shows the body orientation task error for the two task hierarchies. The body orientation performance from $\mathbb{H}_{2}$ is better than that of $\mathbb{H}_{1}$. This can be seen in the interval from 4.0 $\si[per-mode=symbol]{\second}$ to 4.5 $\si[per-mode=symbol]{\second}$ in Fig. \ref{fig:shaking_body} (b). In addition, the different hierarchies cause not only different movements but also different torque profiles. As shown in Fig. \ref{fig:shaking_body} (c), higher torques are needed for $\mathbb{H}_{2}$ than for $\mathbb{H}_{1}$.

\subsection{Dynamic Walking with Directional Change}
\begin{figure}
\centering
\includegraphics[width=0.85\columnwidth]{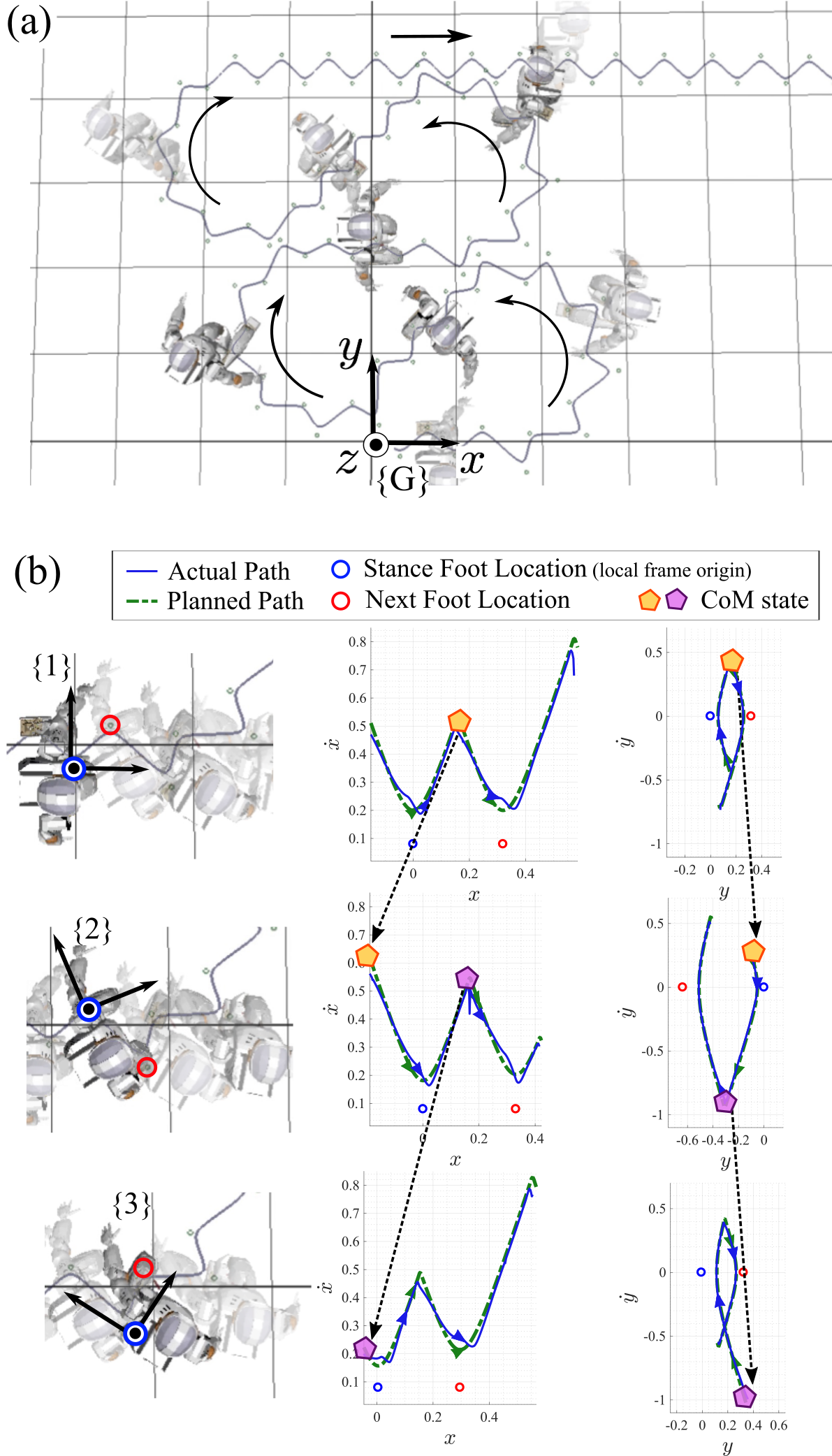}
\caption{{\bf Continuous walking directional change} Valkyrie shows a complex dynamic walking pattern involving changing walking direction. (a) shows a top view of Valkyrie and its walking path. (b) Shos how the robot's CoM state mapped to the next local frame. Local frames rotate with the desired walking direction. For each step, the stance foot becomes the origin of the local frame, and the orientation of the frame is aligned with the desired walking direction. The previous switching CoM state is projected to the current local frame, and the planner finds the foot placement with the new state.}
\label{fig:turn}
\end{figure}

Walking can be broken down into three phases: double contact, right foot contact, and left foot contact. To represent these phases we define the following task hierarchy in WBLC:
\begin{itemize}
\item[$\cdot$] $\ddot{\mathbf{x}}_{1} \in \mathbb{R}^{3}$: Linear CoM position
\item[$\cdot$] $\ddot{\mathbf{x}}_{2} \in \mathbb{R}^{3}$: Pelvis Orientation
\item[$\cdot$] $\ddot{\mathbf{x}}_{3} \in \mathbb{R}^{3}$: Body Orientation
\item[$\cdot$] $\ddot{\mathbf{x}}_{4} \in \mathbb{R}^{3}$: (for the single contact phases) Foot Orientation
\item[$\cdot$] $\ddot{\mathbf{x}}_{5} \in \mathbb{R}^{3}$: (for the single contact phases) Foot Position
\item[$\cdot$] $\ddot{\mathbf{x}}_{6} \in \mathbb{R}^{6}$: Neck and Torso Joint Posture
\item[$\cdot$] $\ddot{\mathbf{x}}_{7} \in \mathbb{R}^{3}$: Centroidal Angular Momentum
\item[$\cdot$] $\ddot{\mathbf{x}}_{8} \in \mathbb{R}^{10}$: Arms Joint Posture 
\end{itemize}
To produce swing foot trajectories, we define third degree B-splines, which guarantee acceleration continuity. The orientation coordinates for the robot's body, pelvis, and feet are described using quaternions. For each step, these orientation tasks are commanded to smoothly switch from the current frame to next one. 
Given initial CoM states, our locomotion planner computes foot positions and their timing while satisfying the desired walking directional changes. In our test shown in Fig. \ref{fig:turn}, Valkyrie takes first 12 steps while continuously changing its walking direction by $18.8^{\si[per-mode=symbol]{\degree}}$ per step. After that, Valkyrie takes 5 forward steps with no directional change. Then, Valkyrie takes another 12 steps while changing direction by, $-18.8^{\si[per-mode=symbol]{\degree}}$ per step. The user only specifies the walking directions while RL-PSP automatically finds the foot positions and their timing using the learned policy. The learned policy consists only on switching states and step locations. The desired position, velocity, and acceleration of the CoM are computed with the analytic equation of the LIP model at runtime.

%%%%%%%%%%%%%%%% Turning Result Figure %%%%%%%%%%%%%%%%%%%%
%%%%%%%%%%%%%%%% End of Turning Result Figure %%%%%%%%%%%%%%%%%%%%
\newcommand*\myswitchyellow{{\protect \includegraphics[width=0.8em]{A_st_state_yellow}}}

\subsection{Push Recovery while Walking}

%%%%%%%%%%%%%%%% Push Recovery Result Figure %%%%%%%%%%%%%%%%%%%%
\begin{figure}
\centering
\includegraphics[width = \columnwidth]{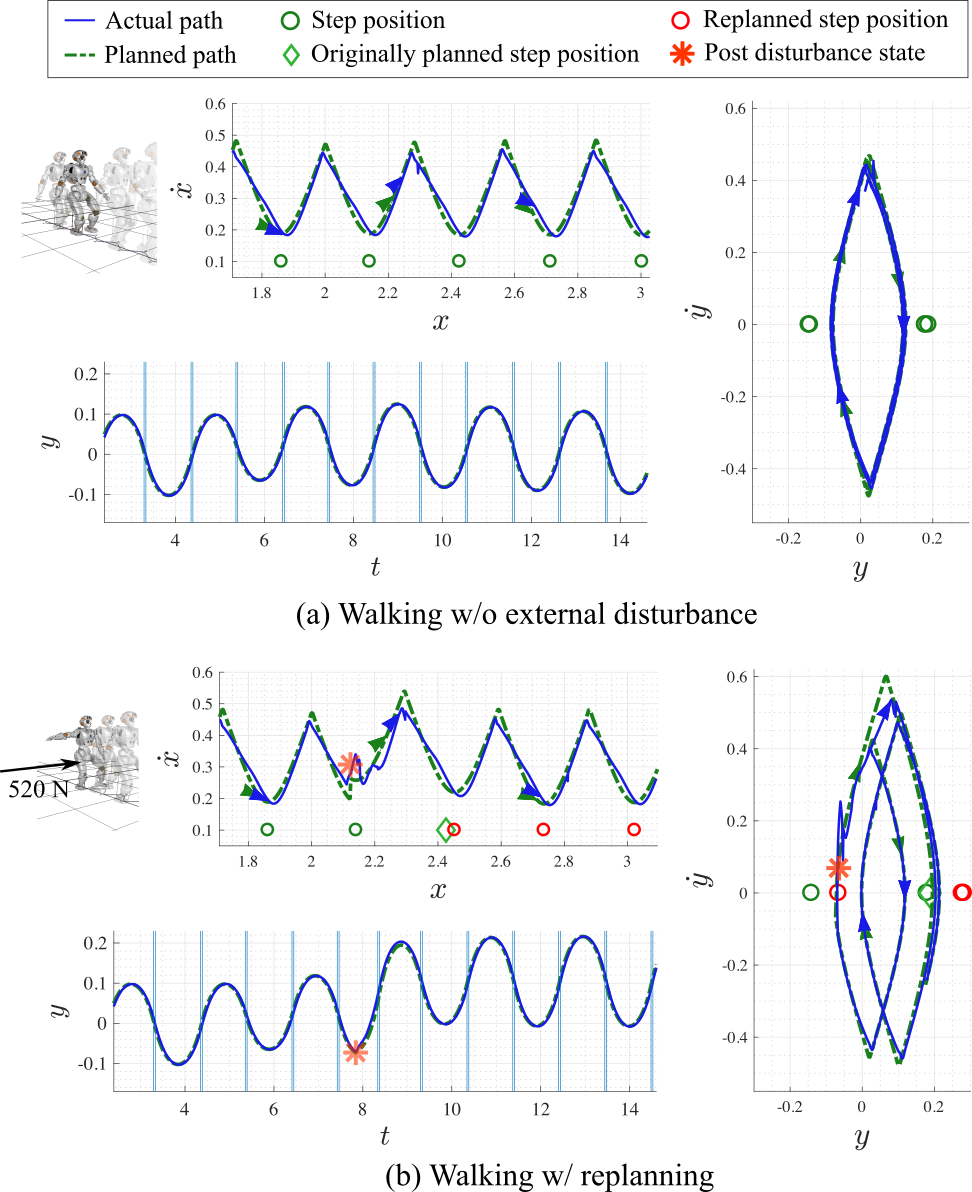}
\caption{{\bf Robustness study.} (a) Shows a walking behavior without external disturbance. (b) When an external impulse of 520 $\si[per-mode=symbol]{\newton}$ and 0.1 $\si[per-mode=symbol]{\second}$ duration is exerted on the robot's pelvis, Valkyrie replans its walking trajectories using the learned policy and maintains its balance without stopping.}
\label{fig:replan}
\end{figure}
%%%%%%%%%%%%%%%% End of Push Recovery Result Figure %%%%%%%%%%%%%%%%%%%%

To validate push recovery, we conduct simulated experiments under large external disturbances and in various directions. Although WBLC is robust to small deviations of the CoM trajectory, for external disturbances we rely on the learned recovery policies described in the theory sections. When the norm of the CoM state error,
\begin{equation}
\mathbf{error}= \begin{bmatrix}
\mathbf{x}^{d} - \mathbf{x}\\
0.5 (\dot{\mathbf{x}}^{d} - \dot{\mathbf{x}})
\end{bmatrix},
\end{equation}
is over a threshold equal to $0.05~\si[per-mode=symbol]{\meter}$ and for longer than 0.02$\si[per-mode=symbol]{\second}$, our planner computes a new trajectory starting from the current CoM state. Instead of setting the new CoM control goal to be the current (disturbed) CoM state, we have experienced that it is better to define a controller goal, $\mathbf{x}^{new}$, equal to 
\begin{equation}
\mathbf{x}^{new} = \gamma\mathbf{x}^{d} + (1-\gamma)\mathbf{x},
\end{equation}
where $\gamma$ can be selected heuristically, and we use a value of 0.8. In our tests, we push Valkyrie while she is dynamically walking using various disturbance forces applied for a duration of 0.1 $\si[per-mode=symbol]{\second}$. The maximum disturbance impulse that we apply to Valkyrie is 520 $\si[per-mode=symbol]{\newton}$ for 0.1 $\si[per-mode=symbol]{\second}$. The results are shown in Fig.~\ref{fig:replan} compared to the undisturbed trajectories. The CoM phase trajectory in the lateral plane shown in Fig.~\ref{fig:replan} (b) shows that the planner is able to find a new trajectory after an external impulse is applied. The time to compute 15 steps after the disturbance is less than 1$\si[per-mode=symbol]{\milli \second}$ using a dual-core 3.0 GHz Intel i7 processor. At the moment that the replanning process occurs, we also find a new swing foot trajectory that transitions from the original swing trajectory to the new goal.

\begin{figure*}
\includegraphics[width=2.0\columnwidth]{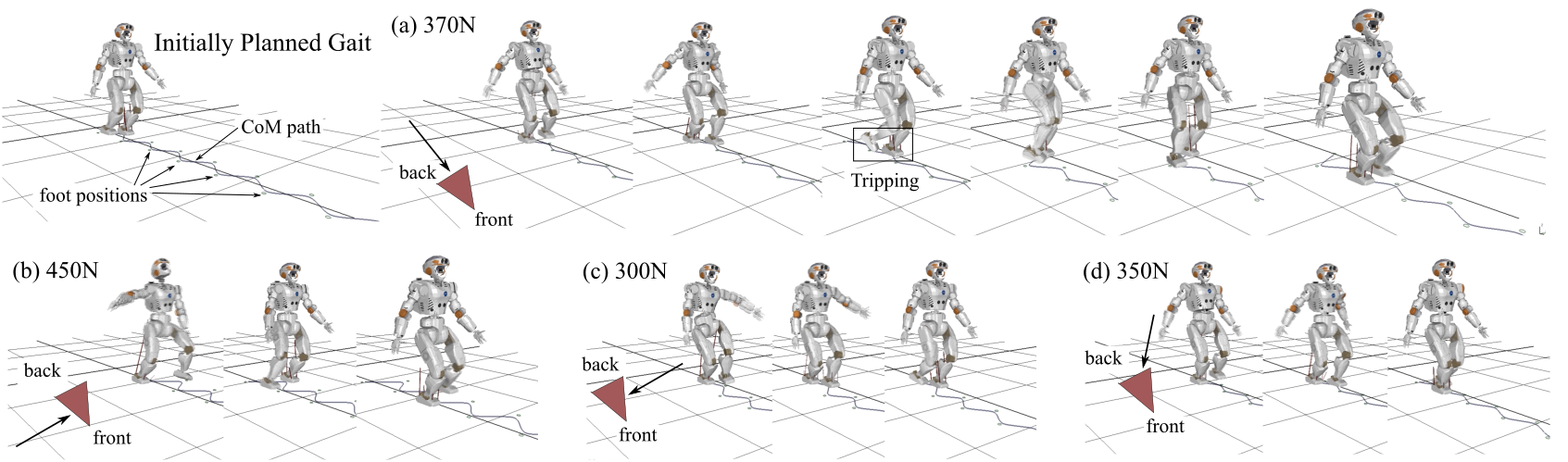}
\caption{{\bf Details of push recovery given various external forces.} In this test we demonstrate the ability of Valkyrie to recover from pushes of various magnitudes and directions. In (a), Valkyrie's feet collide with each other, but the planner finds another path that allows it to recover.}
\label{fig:multi_test}
\end{figure*}

Fig.~\ref{fig:multi_test} shows results of push recoveries while dynamically walking when being subject to various external forces. In all cases, Valkyrie succeeds to sustain the disturbances and continue walking without stopping. The robustness capabilities in this test are competitive to the results of \cite{Khadiv:2017th} which is not based on statistical learning. In contrast to this state of the art, due to our use of offline learning our planner is able to come up with numerous steps almost instantaneously with respect to the walking time frame.

\section{Conclusion and Discussion}
In this paper, we propose an RL based robust locomotion planner and a new WBC, dubbed WBLC. By utilizing PSP in the RL formulation, we can quickly find locomotion policies for 3D walking. The newly developed WBLC takes into account realistic contact and friction cone constraints. At the same time, WBLC maintains task priorities using projection operators which is missing in previous QP based WBCs. Overall, WBLC simultaneously exploits the benefits of QP based WBC's and projection based WBC's, achieving versatility and computational efficiency. Another benefit of our methods is the planning speed. Our locomotion planner almost instantaneously finds a multistep walking trajectory faster than the state of the art. By devising the replanning process during dynamic walking, robots can quickly react to  external forces and achieve significant robustness. 

One interesting aspect of our planning algorithm is the value function we used in the learning process. In the future we could use this value function as an indicator for walking risk given the disturbed states. Many researchers have suggested indicators for locomotion quality. For example, ZMP \cite{VUKOBRATOVIC:2004ej} and CP \cite{Pratt:2006ct} are indicators of balance stability  but they don't take into account other important information such as kinematic constraints or swing time limits. Recently, an allowable CoM acceleration region \cite{Caron:2015cm} has been proposed for multi-contact stability. However, there is no indication of kinematic or dynamic limitations such as step size or swing time. In contrast our value function takes into account some kinematic and dynamic constraints that could ultimately make it a versatile metric for walking quality evaluation.

In the future, we will experiment with more complex functions to represent learned values and policies (e.g. deep neural network). In this paper, we have focused on finding simple walking patterns. However, complex neural networks, which can represent highly nonlinear and abstract behaviors, can enable more versatile planners. For instance, future planners may be able to traverse rough terrain by exploiting various locomotion modes such as walking, running, or jumping. We also plan to implement the proposed algorithms in real systems and evaluate their performance. In our previous work \cite{Kim:2016jg}, we showed agile bipedal balance with a point-foot biped with series elastic actuators. Since the system is highly unstable by nature, we did not apply external disturbances. We believe that the robustness capabilities we have outlined in this paper may allow us ot accomplish sophisticated behaviors in the real testbeds.

\appendices
%\newpage

\section{Analytic Solution of the Phase Space Planner}
\label{sec:append_psp}
When we constraint PIPM dynamics to a piecewise linear height surface, $z = a(x-p_x) + b$, we can find $t_{switch}$ and $p_y$ without numerical integration and bisection search because the system of equations becomes linear, resulting in the following CoM behavior:
\begin{equation}
\label{eq:x_state}
\begin{split}
x(t) & = A e^{\omega t} + B e^{-\omega t} + p_x, \\
\dot{x}(t) & = \omega (A e^{\omega t} - B e^{-\omega t} ),
\end{split}
\end{equation}
where,
\begin{equation}
\begin{split}
\omega &= \sqrt{\frac{g}{a p_x + b}}, \\[2mm]
A &= \frac{1}{2}\Big( (x_{0} - p_x) + \frac{1}{\omega}\dot{x}_{0}   \Big), \\[2mm]
B &= \frac{1}{2}\Big( (x_{0} - p_x) - \frac{1}{\omega}\dot{x}_{0}   \Big).
\end{split}
\end{equation}
Note that this equation is the same for the $y$ direction. Based on Eq.~\eqref{eq:x_state}, we can find an analytical solution for PSP, summarized in Algorithm \ref{code:Phase_Space}. $~\mathbf{x}_{1}$, $\mathbf{y}_{1}$, $\mathbf{x}_{apex,2}$, and $\mathbf{x}_{switch}$ are vector quantities corresponding to the variables $(x_1,\dot{x}_1)$, $(y_1,\dot{y}_1)$, $(x_{apex,2}, \dot{x}_{apex,2})$, and $(x_{switch},\dot{x}_{switch})$.
\begin{algorithm}
\caption{Computation of $t_{switch}$, $p_y$}\label{code:Phase_Space}
% \SetKwFunction{FindXSwitchingState}{Find\_Switching\_State}
% \SetKwFunction{GetTimeAtState}{Get\_Time\_At\_State}
% \SetKwFunction{FindPy}{Find\_Py}
% \SetKwFunction{Integration}{Integration}
\SetKwFunction{FindXSwitchingState}{Find\_Switching\_State}
\SetKwFunction{GetTimeAtState}{Get\_Time}
\SetKwFunction{FindPy}{Find\_Py}
\SetKwFunction{Integration}{GetState}
\KwIn{ $\mathbf{x}_1, \mathbf{y}_1, p_x, \dot{x}_{apex}, \dot{y}_{apex}$}
\KwResult{ $(t_{switch}, p_y)$ } \vspace{1.5mm}
$ \mathbf{x}_{switch} \gets$ \FindXSwitchingState{$\mathbf{x}_1, p_x, \dot{x}_{apex}$} ;\\
\tcp*[f]{Eq.\eqref{eq:vel_x}, \eqref{eq:x_switch}} \vspace{1mm}\\
$ t_{switch} \gets $ \GetTimeAtState{$\mathbf{x}_1, \mathbf{x}_{swtich}$} \tcp*[r]{Eq.\eqref{eq:t_eqn}} \vspace{1.0mm}
$ t_{apex} \gets$ \GetTimeAtState{$\mathbf{x}_{switch}, p_x, \mathbf{x}_{apex}$} \tcp*[r]{Eq.\eqref{eq:t_eqn}} \vspace{1.0 mm}
$\mathbf{y}_{switch} \gets$ \Integration{$\mathbf{y}_1, t_{switch}$}  \tcp*[r]{Eq.\eqref{eq:x_state}} \vspace{1.0 mm}
$p_y$ $\gets$     \FindPy{$\mathbf{y}_{switch}, \dot{y}_{apex}, t_{apex}$} \tcp*[r]{Eq.\eqref{eq:yp}}
\end{algorithm}
Let us focus on obtaining the step switching time. We can easily manipulate Eq.~\eqref{eq:x_state} to analytical solve for the time variable,
\begin{equation}
\begin{split}
& x + \frac{1}{\omega}\dot{x}  = 2A e^{\omega t} + p_x, \\[2mm]
& x + \frac{1}{\omega}\dot{x} - p_x  = 2A e^{\omega t},
\end{split}
\end{equation}
which renders
\begin{equation}
\label{eq:t_eqn}
t  = \frac{1}{\omega} \ln \Big( \frac{x + \frac{1}{\omega}\dot{x} - p_x}{2 A} \Big).
\end{equation}
To find the dynamics, $\dot{x} = f(x)$, which will lead to the switching state solution, let us remove the $t$ term by plugging Eq.~\eqref{eq:t_eqn} into Eq.~\eqref{eq:x_state}.
\begin{equation}
x = A \frac{x + \frac{\dot{x}}{\omega} - p_x}{2 A} + B \frac{2 A}{x + \frac{\dot{x}}{\omega} - p_x } + p_x
\end{equation}
\begin{align}
\frac{1}{2}(x - p_x - \frac{\dot{x}}{\omega}) &=  \frac{2 AB}{x + \frac{\dot{x}}{\omega} - p_x} \\
(x - p_x)^2 - \Big(\frac{\dot{x}}{\omega}\Big)^2  &= 4 AB
\end{align}
By performong some algebra we get, 
\begin{equation}
\begin{split}
\dot{x}^2 &= \omega^2 ( (x - p_x)^2 - 4 AB ),  \\[2mm]
\dot{x}^2 &= \omega^2 \Big( (x - p_x)^2 - (x_{0} - p_x)^2 \Big) + \dot{x}_0^2, 
\end{split}
\end{equation}
which yields,
\begin{equation}
\label{eq:vel_x}
\dot{x} = \pm \sqrt{\frac{g}{h} \Big( (x - p_x)^2 - (x_{0} - p_x)^2 \Big) + \dot{x}_{0}^2 }.
\end{equation}
Given two phase trajectories associate with consecutive walking steps, $p_{x,1}$ and $p_{x,2}$ and assuming the robot walks forward, i.e. $\dot{x}_{switch}$ is positive, we calculate the phase space intersection point via continuity of velocities from Eq.~\eqref{eq:vel_x}:
% The equation gives two $\dot{x}$'s to the $x$, which makes sense because phase plot is mirrored about $x$-axis. However, we can find the switching state where two paths cross because we know the direction of walking, e.g., when robots go forward, $\dot{x}_{switch}$ is always positive. We calculate the intersection point by setting that the velocities (Eq.~\eqref{eq:vel_x} from two different initial states are same, and obtained the following:
%
\begin{equation}
\label{eq:x_switch}
\begin{split}
x_{\rm switch}&=\frac{1}{2}\Big(  \frac{C}{p_{x,2} - p_{x,1}} + (p_{x,1} + p_{x,2}) \Big)\\
C&=(x_{{0},1}-p_{x,1})^2 - (x_{{0},2} - p_{x,2})^2 + \frac{\dot{x}_{{0},2}^2 - \dot{x}_{{0},1}^2}{\omega^2}
\end{split}
\end{equation}
We can now find the step switching time by plugging the computed switching position into Eqs~\eqref{eq:vel_x} and~\eqref{eq:t_eqn}. In addition, we can obtain the timing at the apex velocity from Eq.~\eqref{eq:t_eqn}. The final step is to find the $y$ directional foot placement. We first calculate $\mathbf{y}_{switch}$ by pluggin $t_{switch}$ into the $y$ directional state equation, which has identical form to Eq.~\eqref{eq:x_state}. Then, by using the equality that $\dot{y} (t_{\rm apex})=\dot{y}_{apex}$, we can find $p_{y}$,
\begin{equation}
\label{eq:yp}
\begin{split}
p_y &= \frac{\dot{y}_{apex}-C}{D},\\[2mm]
C &= \frac{\omega}{2} \big( (y_{switch}+\frac{\dot{y}_{switch}}{\omega})e^{\omega t_{apex}} -  \\
&\quad\quad\quad\quad (y_{switch}-\frac{\dot{y}_{switch}}{\omega})e^{-\omega t_{apex}}\big)\\
D &= \frac{\omega}{2}(e^{-\omega t_{apex}} - e^{\omega t_{apex}})
\end{split}
\end{equation}
After calculating $p_y$, we can easily get $y_{apex}$ and $\dot{y}_{apex}$ by using Eq.~\eqref{eq:x_state}.

\section{Equivalent Hierarchy-based Joint Acceleration}
\label{append_b}
The joint velocity associated with an operational task ${\mathbf{x}}_{1}$ is
\begin{equation}
\begin{split}
	\dot{\mathbf{q}} = \bm{J}_{1}^{+} \dot{\mathbf{x}}_{1} + \bm{N}_{1} \dot{\mathbf{q}}_{0}.
\label{eq:qdot_first}
\end{split}
\end{equation}
The definition of the null-space projection matrix using a pseudo inverse and its time derivative yields the following expression:
\begin{equation}
\begin{split}
 \bm{N}_{1} = \bm{I} - \bm{J}_{1}^{+} \bm{J}_{1} &\Rightarrow \dot{\bm{N}}_{1} = -\dot{\bm{J}}_{1}^{+} \bm{J}_{1} - \bm{J}_{1}^{+} \dot{\bm{J}}_{1}.
\end{split}
\end{equation}
The resulting joint acceleration can be obtained by time-derivativating equation (\ref{eq:qdot_first}) as described in \cite{siciliano1991general}
\begin{equation}
\begin{split}
	\ddot{\mathbf{q}}  &= \bm{J}_{1}^{+} \ddot{\mathbf{x}}_{1} +  \dot{\bm{J}}_{1}^{+}\dot{\mathbf{x}}_{1} + \dot{\bm{N}}_{1} \dot{\mathbf{q}}_{0} + \bm{N}_{1}\ddot{\mathbf{q}}_{0} \\
    &= \bm{J}_{1}^{+} \ddot{\mathbf{x}}_{1} + \dot{\bm{J}}_{1}^{+} \bm{J}_{1} \dot{\mathbf{q}} + \dot{\bm{N}}_{1} \dot{\mathbf{q}}_{0}  + \bm{N}_{1}\ddot{\mathbf{q}}_{0} 
\end{split}
\end{equation}
using the equality $\dot{\bm{J}}_{1}^{+} \bm{J}_{1} = - \dot{\bm{N}}_{1} - \bm{J}_{1}^{+} \dot{\bm{J}}_{1}$ we get
\begin{equation}
\begin{split}
    \ddot{\mathbf{q}} &= \bm{J}_{1}^{+} \ddot{\mathbf{x}}_{1} -\bm{J}_{1}^{+} \dot{\bm{J}}_{1} \dot{\mathbf{q}} - \dot{\bm{N}}_{1} \dot{\mathbf{q}} + \dot{\bm{N}}_{1} \dot{\mathbf{q}}_{0}  + \bm{N}_{1}\ddot{\mathbf{q}}_{0} \\
\end{split}
\label{eq:ddotq_1}
\end{equation}
This allows us to simplify equation (\ref{eq:ddotq_1}) to
\begin{equation}
\begin{split}
   \ddot{\mathbf{q}} &= \bm{J}_{1}^{+}\left( \ddot{\mathbf{x}}_{1} - \dot{\bm{J}}_{1} \dot{\mathbf{q}}\right) -\dot{\bm{N}}_{1} \bm{J}_{1}^{+}\dot{\mathbf{x}}_{1} + \bm{N}_{1} \ddot{\mathbf{q}}_0.
\end{split}
\label{eq:qdot_middle}
\end{equation}
If we consider a secondary task $\mathbf{x}_2$, the term $\dot{\mathbf{q}}_0$ becomes
\begin{equation}
	\dot{\mathbf{q}}_{0} = \left( \bm{J}_{2} \bm{N}_{1} \right)^{+} \left( \dot{\mathbf{x}}_{2} - \bm{J}_{2} \bm{J}_{1}^{+} \dot{\mathbf{x}}_{1} \right)
\label{eq:qdot_0}
\end{equation}
because it can be shown that $\bm{N}_{1} \left( \bm{J}_{2} \bm{N}_{1} \right)^{+} = \left( \bm{J}_{2} \bm{N}_{1} \right)^{+}$, we get, 
\begin{equation}
\begin{split}
    \dot{\mathbf{q}}_{0} &= \bm{N}_{1} \dot{\mathbf{q}}_{0} \\
    	\ddot{\mathbf{q}}_0 &= \bm{N}_{1} \ddot{\mathbf{q}}_0 + \dot{\bm{N}}_{1} \dot{\mathbf{q}}_0 \textrm{.}
\label{eq:qdot_nqdot}
\end{split}
\end{equation}
% By multiplying the null-space projection matrix $\bm{N}_{1}$ in equation (\ref{eq:qdot_middle}), we can figure out 
% \begin{equation}
% 	\bm{N}_{1} \ddot{\mathbf{q}} = \bm{N}_{1} \ddot{\mathbf{q}}_{0} - \dot{\bm{N}}_{1} \bm{J}_{1}^{+} \dot{\mathbf{x}}_{1}
% \end{equation}
% because $\bm{N}_{1} \bm{J}_{1}^{+} = \bm{0}$ and $\bm{N}_{1} \dot{\bm{N}}_{1} \bm{J}_{1}^{+} = \dot{\bm{N}}_{1} \bm{J}_{1}^{+}$ 
% And, from equation (\ref{eq:qdot_first}), we can simply obtain     
% \begin{equation}
% \begin{split}
%     \bm{N}_{1} \dot{\mathbf{q}} &= \bm{N}_{1}\dot{\mathbf{q}}_{0} = \dot{\mathbf{q}}_{0} \\
%    \ddot{\mathbf{q}}_{0} &= \dot{\bm{N}}_{1} \dot{\mathbf{q}} + \bm{N}_{1} \ddot{\mathbf{q}} 
% \end{split}
% \end{equation}
From Eq. (\ref{eq:qdot_0}), the term $\ddot{\mathbf{q}}_{0}$ becomes
\begin{equation}
\begin{split}
 \ddot{\mathbf{q}}_{0} &= \bm{J}_{2|1}^{+} \left( \ddot{\mathbf{x}}_{2} - \dot{\bm{J}}_{2} \bm{J}_{1}^{+} \dot{\mathbf{x}}_{1} - \bm{J}_{2} \dot{\bm{J}}_{1}^{+} \dot{\mathbf{x}}_{1} - \bm{J}_{2} \bm{J}_{1} \ddot{\mathbf{x}}_{1} \right) \\
 &+ \dot{\bm{J}}_{2|1}^{+}\left( \dot{\mathbf{x}}_{2} - \bm{J}_{2}\bm{J}_{1}^{+}\dot{\mathbf{x}}_{1} \right) 
\end{split}
\end{equation}
where $ \bm{J}_{2|1}\triangleq \bm{J}_{2} \bm{N}_{1}$ and 
\begin{equation}
	\dot{\bm{J}}_{2|1}^{+} = - \bm{J}_{2|1}^{+} \dot{\bm{J}}_{2|1} \bm{J}_{2|1}^{+}.
\end{equation}
Then, we can manipulate the equation above defining $\ddot{\mathbf{q}}_{0}$ to yield
\begin{equation}
\begin{split}
 \ddot{\mathbf{q}}_{0} &= \bm{J}_{2|1}^{+} \left( \ddot{\mathbf{x}}_{2} - \dot{\bm{J}}_{2} \bm{J}_{1}^{+} \dot{\mathbf{x}}_{1} - \bm{J}_{2} \dot{\bm{J}}_{1}^{+} \dot{\mathbf{x}}_{1} - \bm{J}_{2} \bm{J}_{1}^{+} \ddot{\mathbf{x}}_{1} \right)\\
 & - \bm{J}_{2|1}^{+} \dot{\bm{J}}_{2|1} \dot{\mathbf{q}}_{0}\\
 &= \bm{J}_{2|1}^{+} \left\{ \ddot{\mathbf{x}}_{2} - \dot{\bm{J}}_{2} \dot{\mathbf{q}} - \bm{J}_{2} \bm{J}_{1}^{+} \left(\ddot{\mathbf{x}}_{1} - \dot{\bm{J}}_{1} \dot{\mathbf{q}} \right) \right\} \\
 & + \bm{J}_{2|1}^{+} \left( \dot{\bm{J}}_{2} \dot{\mathbf{q}}_{0} - \bm{J}_{2} \bm{J}_{1}^{+} \dot{\bm{J}}_{1} \dot{\mathbf{q}} - \bm{J}_{2} \dot{\bm{J}}_{1}^{+} \dot{\mathbf{x}}_{1} - \dot{\bm{J}}_{2|1} \dot{\mathbf{q}}_{0} \right)
\end{split}
\end{equation}
For simplicity, we define $ \bm{X}\triangleq\ddot{\mathbf{x}}_{2} - \dot{\bm{J}}_{2} \dot{\mathbf{q}} - \bm{J}_{2} \bm{J}_{1}^{+} \left(\ddot{\mathbf{x}}_{1} - \dot{\bm{J}}_{1} \dot{\mathbf{q}} \right)$. Then the equation on $\ddot{\mathbf{q}}_{0}$ can be further expressed as 
\begin{equation}
\begin{split}
\ddot{\mathbf{q}}_{0} &= \bm{J}_{2|1}^{+} \bm{X}
 + \bm{J}_{2|1}^{+} \left(-\bm{J}_{2} \bm{J}_{1}^{+} \dot{\bm{J}}_{1} \dot{\mathbf{q}} - \bm{J}_{2} \dot{\bm{J}}_{1}^{+} \dot{\mathbf{x}}_{1} - \bm{J}_{2} \dot{\bm{N}}_{1} \dot{\mathbf{q}}_{0} \right) \\
 &= \bm{J}_{2|1}^{+} \bm{X} + \bm{J}_{2|1}^{+} \left\{ -\bm{J}_{2} \bm{J}_{1}^{+} \dot{\bm{J}}_{1} \left( \bm{J}_{1}^{+}\dot{\mathbf{x}}_{1}+ \dot{\mathbf{q}}_{0}   \right) \right. \\
 & \left. - \bm{J}_{2} \dot{\bm{J}}_{1}^{+} \dot{\mathbf{x}}_{1} + \bm{J}_{2}\left(\dot{\bm{J}}_{1}^{+} \bm{J}_{1} + \bm{J}_{1}^{+} \dot{\bm{J}}_{1} \right)  \dot{\mathbf{q}}_{0} \right\} \\
 &= \bm{J}_{2|1}^{+} \bm{X} + \bm{J}_{2|1}^{+} \left(  -\bm{J}_{2} \bm{J}_{1}^{+} \dot{\bm{J}}_{1}  \bm{J}_{1}^{+}\dot{\mathbf{x}}_{1} - \bm{J}_{2} \dot{\bm{J}}_{1}^{+} \dot{\mathbf{x}}_{1} +\bm{J}_{2} \dot{\bm{J}}_{1}^{+} \bm{J}_{1}\dot{\mathbf{q}}_{0} \right)
\end{split}
\end{equation}
Because $\bm{J}_{1} \dot{\mathbf{q}}_{0} = \bm{0}$, the previous equation becomes
\begin{equation}
\begin{split}
\ddot{\mathbf{q}}_{0} &= \bm{J}_{2|1}^{+} \bm{X} + \bm{J}_{2|1}^{+} \left(  -\bm{J}_{2} \bm{J}_{1}^{+} \dot{\bm{J}}_{1}  \bm{J}_{1}^{+}\dot{\mathbf{x}}_{1} - \bm{J}_{2} \dot{\bm{J}}_{1}^{+} \dot{\mathbf{x}}_{1} \right)\\
&= \bm{J}_{2|1}^{+} \bm{X} + \bm{J}_{2|1}^{+} \left( \bm{J}_{2} \bm{J}_{1}^{+} \bm{J}_{1}  \dot{\bm{J}}_{1}^{+}\dot{\mathbf{x}}_{1} - \bm{J}_{2} \dot{\bm{J}}_{1}^{+} \dot{\mathbf{x}}_{1} \right)\\
&=\bm{J}_{2|1}^{+} \bm{X} + \bm{J}_{2|1}^{+}  \bm{J}_{2} \left( \bm{J}_{1}^{+} \bm{J}_{1} - \bm{I}  \right) \dot{\bm{J}}_{1}^{+} \dot{\mathbf{x}}_{1}\\
&=\bm{J}_{2|1}^{+} \bm{X} - \bm{J}_{2|1}^{+}  \bm{J}_{2} \bm{N}_{1} \dot{\bm{J}}_{1}^{+} \dot{\mathbf{x}}_{1}
 \textrm{.}
\end{split}
\end{equation}
Let us develop the term below using the above expression,
\begin{equation}
\begin{split}
-\dot{\bm{N}}_{1} \bm{J}_{1}^{+}\dot{\mathbf{x}}_{1} + \bm{N}_{1} \ddot{\mathbf{q}}_0 &= -\dot{\bm{N}}_{1} \bm{J}_{1}^{+}\dot{\mathbf{x}}_{1} + \ddot{\mathbf{q}}_0 \\
&=\bm{J}_{2|1}^{+} \bm{X} - \dot{\bm{N}}_{1} \bm{J}_{1}^{+}\dot{\mathbf{x}}_{1} - \bm{J}_{2|1}^{+}  \bm{J}_{2} \bm{N}_{1} \dot{\bm{J}}_{1}^{+} \dot{\mathbf{x}}_{1} \\
&= \bm{J}_{2|1}^{+} \bm{X} + \left(\bm{I}- \bm{J}_{2|1}^{+}  \bm{J}_{2} \bm{N}_{1} \right) \bm{N}_{1} \dot{\bm{J}}_{1}^{+} \dot{\mathbf{x}}_{1} \\
&= \bm{J}_{2|1}^{+} \bm{X} + \bm{N}_{2|1} \bm{N}_{1} \dot{\bm{J}}_{1}^{+} \dot{\mathbf{x}}_{1} 
\end{split}
\end{equation}
Thus, equation (\ref{eq:qdot_middle}) becomes
\begin{equation}
\begin{split}
   \ddot{\mathbf{q}} &= \bm{J}_{1}^{+}\left( \ddot{\mathbf{x}}_{1} - \dot{\bm{J}}_{1} \dot{\mathbf{q}}\right) + \bm{J}_{2|1}^{+} \bm{X} + \bm{N}_{2|1} \bm{N}_{1} \dot{\bm{J}}_{1}^{+} \dot{\mathbf{x}}_{1}\\
   &=  \bm{J}_{1}^{+}\left( \ddot{\mathbf{x}}_{1} - \dot{\bm{J}}_{1} \dot{\mathbf{q}}\right) + \bm{J}_{2|1}^{+} \bm{X} + \bm{N}_{2|1} \bm{N}_{1} \ddot{\mathbf{q}}_{res}
\end{split}
\label{eq:qdot_final}
\end{equation}

% This relationship is exactly equivalent the formulation used in this paper
% \begin{equation}
% 	\ddot{\mathbf{x}}_{1} = \bm{J}_{1} \ddot{\mathbf{q}} + \dot{\bm{J}}_{1} \dot{\mathbf{q}} 
% \end{equation}
% and
% \begin{equation}
% \begin{split}
% 	&\ddot{\mathbf{q}} = \bm{J}_{1}^{+} \left( \ddot{\mathbf{x}}_{1} - \dot{\bm{J}}_{1} \dot{\mathbf{q}} \right) \\
%     \Rightarrow &\quad \ddot{\mathbf{q}} = \bm{J}_{1}^{+}\left( \ddot{\mathbf{x}}_{1} - \dot{\bm{J}}_{1} \dot{\mathbf{q}} \right) + \bm{N}_{1}\widehat{\ddot{\mathbf{q}}}_{0}
% \end{split}    
% \end{equation}
% where 
% \begin{equation}
% \widehat{\ddot{\mathbf{q}}}_{0} = \bm{J}_{2|1}^{+} \bm{X} \textrm{.}
% \end{equation}
% Then, we can extend the formulation for the residual joint acceleration described in this paper. 
% \begin{equation}
% \begin{split}
% \ddot{\mathbf{q}} &= \bm{J}_{1}^{+}\left( \ddot{\mathbf{x}}_{1} - \dot{\bm{J}}_{1} \dot{\mathbf{q}} \right) + \bm{N}_{1}\widehat{\ddot{\mathbf{q}}}_{0} +\overrightarrow{\bm{N}_{2}} \widehat{\ddot{\mathbf{q}}}_{res} \\
% &= \bm{J}_{1}^{+}\left( \ddot{\mathbf{x}}_{1} - \dot{\bm{J}}_{1} \dot{\mathbf{q}} \right) + \bm{J}_{2|1}^{+} \bm{X} + \bm{N}_{2|1} \bm{N}_{1} \widehat{\ddot{\mathbf{q}}}_{res}
% \end{split}
% \end{equation}
% which is exactly same to Equation (\ref{eq:qdot_final}).

%%%%%%%%%%%%%%%%%%%%%%%%%%%%%%%%%%%%%%%%%%%%%%%%%%%%%%%%%%
%%%                   Related Work                      %%
%%%%%%%%%%%%%%%%%%%%%%%%%%%%%%%%%%%%%%%%%%%%%%%%%%%%%%%%%%
\section{Related Work}
\label{sec:related_work}

%%%%%% RL - Locomotion %%%%
\subsection{Reinforcement Learning based Locomotion Planner}
One of the main challenges for learning robust dynamic locomotion policies is  handling the high number of continuous variables describing the motion and force interactions of full humanoid robots. To deal with the dimensionality problem, we review previous work that has greatly inspired us. \cite{Morimoto:2007eh}, solves a periodic locomotion generation problem via RL on a planar biped robot. We advance upon this work by solving the learning problem for 3D robots, avoiding reliance on human walking trajectories, and generating policies for non-periodic gaits. Other important works employ learning as an optimization problem over known locomotion trajectories. In \cite{Sugimoto:2011ge}, a periodic locomotion problems is solved by optimizations of a known stable central patter generated (CPG) walking trajectories using RL. However, no focus is given to dealing with large external disturbances. In addition, our focus is on the generation of trajectories from scratch without prior stable locomotion patterns.  

In \cite{Missura:2014kn}, robust walking trajectories to external pushes are achieved based on capture point trajectory optimization via gradient based learning updates. In this work, the capture point method is used as an analytic controller to initiate the learning process with information about foot placement, step-timing, and ZMP controls.  Although the authors also show learning of push recovery strategies without previous capture point generated trajectories, our focus is stronger on autonomously learning the locomotion process without reliance on already stable walking gaits. As such, we belief our algorithm is able to learn from scratch recovery strategies in a more generic sense, for instance to recover from pushes in any direction while walking. Like ours, autonomous learning of periodic gaits has been explored before in passive dynamic walkers \cite{Tedrake:2004ip}. Once more, our focus is on gait generators that can produce non-periodic gaits and tolerate large push disturbances in all directions of motion.
% \cite{MacAlpine:2012vp} optimized the parameters of walking engines. As the algorithms are based on a few specific robots or walking engines, applying these methods to other systems is difficult. 

The dynamic locomotion community has previously used online optimization methods instead of RL, such as model predictive control (MPC). The main problem of these approaches is the high computational cost. To mitigate this problem, researchers have made significant efforts to develop efficient computational processes. \cite{Erez:2013cl} used the gradient of a cost function to solve the MPC problem efficiently. \cite{Khadiv:2017th} linearized the planning problem by optimizing over one step ahead of time. Our approach relying on learned neural networks replaces the need for complex online computations, enabling the generation of hundreds of steps in an instant compared to the stepping time scales.
% Optimization-based algorithms share a few characteristics such as linearization, carefully adjusted time horizon,  and relatively expensive computation. However, we can replace the online optimization process by training neural networks to learn the optimal policy offline. 
\cite{Whitman:2009im} proposed a controller for a 12 DoF biped system by using dynamic programming and a lookup table that was obtained offline based on simple models. The multiple policies achieved from each simple model were combined to control the target system. In contrast, our work relies on the generic inverted pendulum locomotion model, and a versatile full-humanoid body controller, i.e., WBLC. 
% Moreover, we prevent the problem of dimensionality by learning a policy through RL instead of writing a lookup table through dynamic programming. Therefore, the proposed algorithm has the benefit that we can easily apply the proposed method to other systems and demonstrate robust biped walking.

\subsection{Whole-Body Control}
WBC \cite{sentis2005synthesis} is a family of multi and prioritized task-space trajectory controllers for humanoid robots that rely on floating-base dynamic and computed torque commands as inputs to the plant. It yields asymptotically stable control policies for multiple tasks with simultaneous control of operational forces when needed. Priorities address resource allocation when two or more task trajectories cannot physically be tracked by the robotic system. It naturally integrates equality constraints such as biarticular transmission constraints \cite{sentis2013implementation}. 
Other groups have explored richer versions of WBC with inequality constraints such as  joint limit avoidance \cite{flacco2012motion,mansard2009unified,lee2012intermediate}, collision avoidance \cite{kanoun2011kinematic}, and singularity avoidance \cite{moe2015stability}. Several groups used evolved and more practical versions of WBC such as controllers used in the DARPA Robotics Challenge of 2013 and 2014. For instance, \cite{koolen2013summary, johnson2015team} incorporate reaction forces as inequality constraints based on solving a quadratic programming optimization problem with desired center of mass trajectories. Treatment of reaction forces as inequality constraints in the WBC communities dates back to the work by \cite{stephens2010dynamic}. And it showcases one of the weaknesses of our group's formulation of WBC. In early versions \cite{sentis2010compliant} we treated reaction forces as equality constraints. Such treatment corresponds to bilateral contact constraints, i.e. assuming that the floor contacts are actually rigid anchors. This is obviously an inaccurate model. One of the main objectives of this paper is to use a realistic unilateral contact model for WBC while maintaining one of its main strengths, efficient prioritized control.

% Application of WBCs: Walking motion, Locomotion
Bipedal and quadrupedal walking capabilities have been devised using WBC. \cite{hutter2014quadrupedal} demonstrated locomotion of a quadrupedal robot by utilizing hierarchical tasks based on least-square problems. The integration of the versatile capture point (CP) as an operational space of WBC was proposed and controlled either as a constraint or a task for bipedal humanoid robots \cite{ramos2014whole}. The robot's Center of Gravity (CoG) has been used as a task controller for a while, such as in \cite{mistry2007task}. Walking pattern generators have been incorporated into WBC in multiple instances such as in \cite{carpentier2016versatile}. During the DARPA robotics challenge, several top participants incorporated WBC's into their strategy for achieving mobile dexterous capabilities. For instance, high-level trajectory optimization and low-level optimization with inverse dynamics were integrated into the framework by \cite{feng2015optimization}. 

% Optimization-based WBCs
As stated before, during the DRC several humanoid robots were controlled via WBC including QP solvers for dealing with reaction forces. By introducing QP and task hierarchy (HQP), whole-body motion of humanoid robots could be controlled with the intrinsic reactive advantages of task prioritization \cite{escande2014hierarchical}. Compared with projection-based WBC algorithms, optimization-based WBC, such as HQP, can incorporate multiple inequality constraints \cite{saab2013dynamic}, which are useful for describing contact conditions such as friction cones \cite{abe2007multiobjective}. Overall, optimization-based WBC have been a success for practical applications \cite{koolen2013summary,feng2015optimization,kuindersma2014efficiently}. However, their computational cost remains a challenge, specially if being considered as models for motion planning, such as model predictive control. Therefore, efficiency of our newly proposed whole-body controller, dubbed WBLC, is a key consideration of this paper. To achieve the speed boost, we rely on a projection-base formulation. However, it is difficult to incorporate inequality constraints into analytical projection-based methods; thus, our goal is to combine both and also to maintain desired task hierarchy capabilities. The proposed WBLC incorporates an efficient QP, the dimension of which depends only on the number of contact points, and a joint acceleration level controller which only relies on projective operators, thus yielding the speed efficiency that we advocate for.

\section*{Acknowledgment}
The authors would like to thank the members of the Human Centered Robotics Laboratory at The University of Texas at Austin for their great help and support. This work was supported by the Office of Naval Research, ONR Grant [grant \#N000141512507] and NASA Johnson Space Center, NSF/NASA NRI Grant [grant \#NNX12AM03G].
\ifCLASSOPTIONcaptionsoff
  \newpage
\fi

\bibliographystyle{IEEEtran}
\bibliography{tro2017}

%%%%%%%% PROFILE %%%%%%%%%%%%
% \input{profile.tex}

\end{document}